\documentclass{article} %
\usepackage{iclr2023_conference,times}

\usepackage[utf8]{inputenc} %

\usepackage{graphicx}

\usepackage[T1]{fontenc}

\IfFileExists{headers/config/showoverfull.config}{
	\overfullrule=1cm
}{
}

\usepackage{marginnote}

\usepackage[backgroundcolor=none,linecolor=red,textsize=footnotesize]{todonotes}

\usepackage{etoolbox}

\newbool{includeappendix}
\setbool{includeappendix}{true} %
\IfFileExists{headers/config/noappendix.config}{
	\setbool{includeappendix}{false}
}{}

\newif\ifincludeappendixx
\ifbool{includeappendix}{
	\includeappendixxtrue
}{
	\includeappendixxfalse
}

\usepackage{xr} %
\usepackage{filecontents}

\ifbool{includeappendix}{}{
	\input{appendix-labels-loader}

	\externaldocument{appendix-labels}
}

\usepackage{subcaption}

\definecolor{my-full-blue}{HTML}{1F77B4}

\definecolor{my-full-orange}{HTML}{FF7F0E}

\definecolor{my-full-green}{HTML}{2CA02C}

\definecolor{my-full-red}{HTML}{d62728}

\definecolor{my-full-purple}{HTML}{9467bd}

\definecolor{my-full-brown}{HTML}{8c564b}

\definecolor{my-full-pink}{HTML}{e377c2}

\definecolor{my-full-gray}{HTML}{7f7f7f}

\definecolor{my-full-olive}{HTML}{bcbd22}

\definecolor{my-full-cyan}{HTML}{17becf}

\colorlet{my-blue}{my-full-blue!30}
\colorlet{my-orange}{my-full-orange!30}
\colorlet{my-green}{my-full-green!30}
\colorlet{my-red}{my-full-red!30}
\colorlet{my-purple}{my-full-purple!30}
\colorlet{my-brown}{my-full-brown!30}
\colorlet{my-pink}{my-full-pink!30}
\colorlet{my-gray}{my-full-gray!30}
\colorlet{my-olive}{my-full-olive!30}
\colorlet{my-cyan}{my-full-cyan!30}

\definecolor{fgray}{RGB}{205, 205, 205}
\definecolor{fred}{RGB}{152, 68, 100}
\definecolor{fteal}{RGB}{94, 204, 171}
\definecolor{fpurple}{RGB}{192, 175, 251}
\definecolor{forange}{RGB}{230, 161, 118}
\definecolor{fblue}{RGB}{0, 103, 138}

\usepackage{listings}

\usepackage{textcomp}

\usepackage{xcolor}

\usepackage[scaled=0.8]{beramono}

\definecolor{ckeyword}{HTML}{7F0055}
\definecolor{ccomment}{HTML}{3F7F5F}
\definecolor{cstring}{HTML}{2A0099}

\lstdefinestyle{numbers}{
	numbers=left,
	framexleftmargin=20pt,
	numberstyle=\tiny,
	firstnumber=auto,
	numbersep=1em,
	xleftmargin=2em
}

\lstdefinestyle{layout}{
	frame=none,
	captionpos=b,
}

\lstdefinestyle{comment-style}{
	morecomment=[l]//,
	morecomment=[s]{/*}{*/},
	commentstyle={\color{ccomment}\itshape},
}

\lstdefinestyle{string-style}{
	morestring=[b]",%
	morestring=[b]',%
	stringstyle={\color{cstring}},
	showstringspaces=false,%
}

\lstdefinestyle{keyword-style}{
	keywordstyle={\ttfamily\bfseries},
	morekeywords={
		function,
		constructor,
		int,
		bool,
		return,
		returns,
		uint
	},
	morekeywords = [2]{},
	keywordstyle = [2]{\text},
	sensitive=true,
}

\lstdefinestyle{input-encoding}{
	inputencoding=utf8,
	extendedchars=true,
	literate=
	{ℝ}{$\reals$}1%
	{→}{$\rightarrow$}1%
	{α}{$\alpha$}1%
	{β}{$\beta$}1%
	{λ}{$\lambda$}1%
	{θ}{$\theta$}1%
	{ϕ}{$\phi$}1%
}

\lstdefinestyle{escaping}{
	moredelim={**[is][\color{blue}]{\%}{\%}},
	escapechar=|,
	mathescape=true
}

\lstdefinestyle{default-style}{
	basicstyle=\fontencoding{T1}\ttfamily\footnotesize,
	style=numbers,
	style=layout,
	style=comment-style,
	style=string-style,
	style=keyword-style,
	style=input-encoding,
	style=escaping,
	tabsize=2,
	upquote=true
}

\lstdefinelanguage{BASIC}{
	language=C++,
	style=default-style
}[keywords,comments,strings]%

\usepackage{tikz}

\usetikzlibrary{arrows}
\usetikzlibrary{arrows.meta}
\usetikzlibrary{automata}
\usetikzlibrary{calc}
\usetikzlibrary{backgrounds}
\usetikzlibrary{decorations}
\usetikzlibrary{decorations.markings}
\usetikzlibrary{decorations.pathmorphing}
\usetikzlibrary{decorations.pathreplacing}
\usetikzlibrary{fit}
\usetikzlibrary{intersections}
\usetikzlibrary{patterns}
\usetikzlibrary{positioning}
\usetikzlibrary{shadows}
\usetikzlibrary{shapes}
\usetikzlibrary{shapes.geometric}
\usetikzlibrary{shadows,patterns,snakes}
\usetikzlibrary{arrows.meta}
\tikzset{>={Latex[width=2mm,length=3mm]}}
\tikzstyle{terminator} = [rectangle, draw, align = left, rounded corners, minimum height=2em]
\tikzstyle{process} = [rectangle, draw, align = left, minimum height=2em]
\tikzstyle{decision} = [diamond, draw, align = left, minimum height=2em]
\tikzstyle{connector} = [draw, -latex']

\usepackage{pgfplots}
\pgfplotsset{compat=newest}

\usepackage{amsmath,amsfonts,bm}

\def\eqref#1{equation~\ref{#1}}

\def\1{\bm{1}}

\newcommand{\valid}{\mathcal{D_{\mathrm{valid}}}}

\def\vmu{{\bm{\mu}}}

\def\va{{\bm{a}}}
\def\vb{{\bm{b}}}
\def\vc{{\bm{c}}}
\def\vd{{\bm{d}}}
\def\ve{{\bm{e}}}
\def\vf{{\bm{f}}}
\def\vg{{\bm{g}}}

\def\vl{{\bm{l}}}

\def\vs{{\bm{s}}}

\def\vu{{\bm{u}}}

\def\vx{{\bm{x}}}
\def\vy{{\bm{y}}}
\def\vz{{\bm{z}}}

\def\mA{{\bm{A}}}

\def\mW{{\bm{W}}}

\DeclareMathAlphabet{\mathsfit}{\encodingdefault}{\sfdefault}{m}{sl}
\SetMathAlphabet{\mathsfit}{bold}{\encodingdefault}{\sfdefault}{bx}{n}

\newcommand{\R}{\mathbb{R}}

\newcommand{\KL}{D_{\mathrm{KL}}}

\DeclareMathOperator*{\argmax}{arg\,max}
\DeclareMathOperator*{\argmin}{arg\,min}

\DeclareMathOperator{\sign}{sign}

\usepackage{hyperref}
\usepackage{url}

\usepackage{adjustbox}
\usepackage{amsmath,amssymb,amsfonts,mathrsfs}
\usepackage{dsfont}

\usepackage{enumitem}
\usepackage{threeparttable}
\usepackage{multirow, makecell}
\usepackage{booktabs}
\usepackage{xspace}
\usepackage{wrapfig}
\usepackage{xfrac}

\usepackage{subcaption}
\usepackage{caption}
\newcommand{\bc}[1]{\mathcal{#1}}
\newcommand{\bs}[1]{\boldsymbol{#1}}

\DeclareMathOperator*{\csim}{sim}
\DeclareMathOperator*{\ELBO}{ELBO}
\DeclareMathOperator*{\vol}{vol}

\DeclareMathOperator*{\MAE}{MAE}
\DeclareMathOperator*{\MAER}{MAE_\text{rob}}

\DeclareFontFamily{U}{mathx}{\hyphenchar\font45}
\DeclareFontShape{U}{mathx}{m}{n}{
<-6> mathx5 <6-7> mathx6 <7-8> matha7
<8-9> mathx8 <9-10> mathx9
<10-12> mathx10 <12-> mathx12
}{}
\DeclareSymbolFont{mathx}{U}{mathx}{m}{n}
\DeclareMathSymbol{\bigtimes}{\mathop}{mathx}{"91}

\newcommand{\atol}{\tau}

\newcommand{\tool}{\textsc{GAINS}\xspace}
\newcommand{\tooll}{\textbf{G}raph based \textbf{A}bstract \textbf{I}nterpretation for \textbf{N}ODE\textbf{s}\xspace}

\newcommand{\toolt}{\textsc{GAINS}\xspace}

\newcommand{\solver}{\textsc{CAS}\xspace}

\newcommand{\lcap}{LCAP\xspace}
\newcommand{\lcapm}{\textsc{CURLS}\xspace}
\newcommand{\lcapml}{\textbf{C}onstraint \textbf{U}nification via \textbf{R}e\textbf{L}U \textbf{S}implification\xspace}

\newcommand{\deeppoly}{\textsc{DeepPoly}\xspace}

\newcommand{\boxd}{\textsc{Box}\xspace}

\newcommand{\cifar}{CIFAR-10\xspace}

\newcommand{\mnist}{\textsc{MNIST}\xspace}
\newcommand{\fmnist}{\textsc{FMNIST}\xspace}
\newcommand{\physio}{\textsc{Physio-Net}\xspace}

\newcommand{\ode}{\text{ODE}\xspace}
\newcommand{\odes}{\text{ODEs}\xspace}

\newcommand{\node}{\text{NODE}\xspace}
\newcommand{\nodes}{\text{NODEs}\xspace}

\newcommand{\NN}{\text{NN}\xspace}

\newcommand{\relu}{\text{ReLU}\xspace}

\newcommand{\RNN}{\text{RNN}\xspace}

\newcommand{\mae}{\text{MAE}\xspace}

\newcommand{\markerc}[1]{\tikz[]{\node[fill, circle, aspect=1, color=#1, inner sep=0pt, minimum size=2.5mm]{};}\xspace}
\newcommand{\markerb}[1]{\tikz[]{\node[fill, aspect=1, color=#1, inner sep=0pt, minimum size=2.1mm]{};}\xspace}
\newcommand{\ccc}[1]{{\color{black}#1}}

\usepackage[capitalize]{cleveref}

\makeatletter
\newcommand\theHALG@line{\thealgorithm.\arabic{ALG@line}}
\makeatother

\crefformat{section}{\S#2#1#3}

\crefrangeformat{section}{\S#3#1#4\crefrangeconjunction\S#5#2#6}

\crefmultiformat{section}{\S#2#1#3}{\crefpairconjunction\S#2#1#3}{\crefmiddleconjunction\S#2#1#3}{\creflastconjunction\S#2#1#3}

\newcommand{\crefrangeconjunction}{--}

\crefname{listing}{Lst.}{listings}
\crefname{line}{Lin.}{Lin.}
\crefname{appendix}{App.}{App.}

\newcommand{\appref}[1]{%
	\ifbool{includeappendix}{\cref{#1}}{the appendix}%
}
\newcommand{\Appref}[1]{%
	\ifbool{includeappendix}{\cref{#1}}{The appendix}%
}

\title{Efficient Certified Training and Robustness Verification of Neural ODEs}
\author{Mustafa Zeqiri, Mark Niklas Müller, Marc Fischer \& Martin Vechev\\
	ETH Zurich\\
	Zurich, Switzerland\\
	\texttt{mzeqiri@ethz.ch},
	\texttt{\{mark.mueller,mark.fischer,martin.vechev\}@inf.ethz.ch} \\
}

\iclrfinalcopy %

\begin{document}

\maketitle

\begin{abstract}
Neural Ordinary Differential Equations (NODEs) are a novel neural architecture, built around initial value problems with learned dynamics which are solved during inference. Thought to be inherently more robust against adversarial perturbations, they were recently shown to be vulnerable to strong adversarial attacks, highlighting the need for formal guarantees.  However, despite significant progress in robustness verification for standard feed-forward architectures, the verification of high dimensional NODEs remains an open problem. In this work, we address this challenge and propose \tool, an analysis framework for NODEs combining three key ideas: (i) a novel class of ODE solvers, based on variable but discrete time steps, (ii) an efficient graph representation of solver trajectories, and (iii) a novel abstraction algorithm operating on this graph representation. Together, these advances enable the efficient analysis and certified training of high-dimensional NODEs, by reducing the runtime from an intractable $\bc{O}(\exp(d)+\exp(T))$ to $\bc{O}(d+T^2 \log^2T)$ in the dimensionality $d$ and integration time $T$.  In an extensive evaluation on computer vision (\mnist and \fmnist) and time-series forecasting (\physio) problems, we demonstrate the effectiveness of both our certified training and verification methods.
\end{abstract}

\section{Introduction} \label{sec:introduction}

As deep learning-enabled systems are increasingly deployed in safety-critical domains, developing neural architectures and specialized training methods that increase their robustness against adversarial examples \citep{szegedy2013intriguing,BiggioCMNSLGR13} -- imperceptible input perturbations, causing model failures -- is more important than ever. 
As standard neural networks suffer from severely reduced accuracies when trained for robustness, novel architectures with inherent robustness properties have recently received increasing attention \citep{WinstonK20,MuellerSFV21}.

\paragraph{Neural Ordinary Differential Equations} One particularly interesting such architecture are neural \odes (\nodes) \citep{chen2018neural}. Built around solving initial value problems with learned dynamics, they are uniquely suited to time-series-based problems \citep{rubanova2019latent,de2019gru} but have also been successfully applied to image classification \citep{chen2018neural}. 
More importantly, \nodes have been observed to exhibit inherent robustness properties against adversarial attacks \citep{yan2019robustness,kang2021stable,rodriguez2022lyanet,zakwan2022robust}. 
However, recently \citet{huang2020adversarial} found that this robustness is greatly diminished against stronger attacks. They suggest that adaptive \ode solvers, used to solve the underlying initial value problems, cause gradient obfuscation \citep{athalye2018obfuscated}, which, in turn, causes weaker adversarial attacks to fail. This highlights the need for formal robustness guarantees to rigorously evaluate the true robustness of a model or architecture.

\paragraph{Robustness Verification} For standard neural networks, many robustness verification methods have been proposed \citep{KatzBDJK17,TjengXT19,singh2018fast,RaghunathanSL18,WangZXLJHK21,FerrariMJV22}. One particularly successful class of such methods \citep{GehrMDTCV18,singh2019abstract,zhang2018crown} propagates convex shapes through the neural network that capture the reachable sets of every neuron's values and uses them to check whether a given robustness property holds.
Unfortunately, none of these methods can be applied to \nodes as the underlying adaptive solvers yield a continuous range of possible step-sizes (illustrated in the top panel of \cref{fig:intro}), which existing analysis techniques can not handle.
First works towards \node verification \citep{lopez2022reachability}  have avoided this issue by disregarding the solver behavior and analyzing only the underlying \node dynamics in extremely low dimensional settings. 
However, both scaling to high-dimensional \node architectures and taking the effect of \ode solvers into account remain open problems preventing \node robustness verification.

\input{figures/intro}

\paragraph{This Work} We tackle both of these problems, thereby enabling the systematic verification and study of \node robustness (illustrated in \cref{fig:intro}) as follows:
(i) We introduce a novel class of \ode solvers, based on the key insight that we can restrict step-sizes to an exponentially spaced grid with minimal impact on solver efficiency, while obtaining a finite number of time/step-size trajectories from the initial to final state (see the second column in \cref{fig:intro}). We call these \textbf{C}ontrolled \textbf{A}daptive \ode \textbf{S}olvers (\solver). 
Unfortunately, \solver solvers still yield exponentially many trajectories in the integration time. (ii) We, therefore, introduce an efficient graph representation, allowing trajectories to be merged, reducing their number to quadratically many.
(iii) We develop a novel algorithm for the popular \deeppoly convex relaxation \citep{singh2019abstract}, effective for standard neural network verification, that enables the efficient application of \deeppoly to the trajectory graph by handling trajectory splitting in linear instead of exponential time.
Combining these core ideas, we propose \tool\footnote{\tooll}, a novel framework for certified training and verification of \nodes that leverages key algorithmic advances to achieve polynomial runtimes and allows us to faithfully assess the robustness of \nodes.

\paragraph{Main Contributions} Our main contributions are:
\begin{itemize}
    \item A novel class of \ode solvers, \solver solvers, retaining the efficiency of adaptive step size solvers while enabling verification (\cref{sec:controlled-adaptive}).
    \item An efficient linear bound propagation based framework, \tool, which leverages \solver to enable the efficient training and verification of \nodes (\cref{sec:verification}).
    \item An extensive empirical evaluation demonstrating the effectiveness of our method in ablation studies and on image classification and time-series problems (\cref{sec:experiments}).
\end{itemize}
\newpage

\section{Adversarial Robustness} \label{sec:background}
In this section, we discuss the necessary background relating to adversarial robustness.%

\paragraph{Adversarial Robustness}
We consider both classification and regression models $\vf_{\bs{\theta}} \colon \R^{d_\text{in}} \mapsto \R^{c}$ with parameters $\bs{\theta}$ that, given an input $\vx \in \bc{X} \subseteq \R^{d_\text{in}}$, predict $c$ numerical values $\vy := \vf(\vx)$, interpreted as class confidences or predictions of the regression values, respectively. 
In the classification setting, we call $\vf$ adversarially robust on an $\ell_p$-norm ball $\bc{B}_p^{\epsilon_p}(\vx)$ of radius $\epsilon_p$, if it predicts target class $t$ for all perturbed inputs $\vx' \in \bc{B}_p^{\epsilon_p}(\vx)$. More formally, we define \emph{adversarial robustness} as:
\begin{equation}
\label{eq:adv_robustness_cls}
\argmax_j h(\vx')_j = t, \quad \forall \vx' \in \bc{B}_p^{\epsilon_p}(\vx) := \{x' \in \bc{X} \mid \|\vx -\vx'\|_p \leq \epsilon_p\}.
\end{equation}
In the regression setting, we evaluate $\nu$-$\delta$-robustness by checking whether the worst-case mean absolute error $\MAER(\vx)$ for $\vx' \in \bc{B}_p^{\epsilon_p}(\vx)$ is linearly bounded by the original input's $\MAE(\vx)$:
\begin{equation}
\label{eq:adv_robustness_reg}
\MAER(\vx) < (1 + \nu) \MAE(\vx) + \delta, \quad \text{with} \quad \MAER(\vx) = \max_{\vx' \in \bc{B}_p^{\epsilon_p}(\vx)} \MAE(\vx').
\end{equation}

\paragraph{Adversarial Attacks} aim to disprove robustness properties by finding a concrete counterexample $\vx'$. A particularly successful such method is the PGD attack \citep{madry2017towards}, which computes $\vx'$ by initializing $\vx'_0$ uniformly at random in $\bc{B}_p^{\epsilon_p}(\vx)$ and then updating it in the direction of the gradient sign of an auxiliary loss function $\bc{L}$, using $N$ projected gradient descent steps:
\begin{equation}
\vx'_{n+1}=\Pi_{\bc{B}_p^{\epsilon_p}(\vx)}\vx'_n + \alpha \sign(\nabla_{\vx'_n} \bc{L}(\vf_{\bs{\theta}}(\vx'_n),t)),
\end{equation}
where $\Pi_S$ denotes projection on $S$ and $\alpha$ the step size.
We say an input $\vx$ is empirically robust if no counterexample $\vx'$ is found.

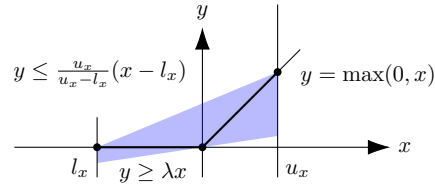
\begin{wrapfigure}[8]{r}{0.45\textwidth}
\vspace{-6mm}
    \centering
    \begin{tikzpicture}
    	\draw[->] (-2.5, 0) -- (2.5, 0) node[right,scale=0.85] {$x$};
    	\draw[->] (0, -0.4) -- (0, 1.6) node[above,scale=0.85] {$y$};

    	\def\a{-1.4}
    	\def\b{1.0}
    	\def\al{0.15}
    	\coordinate (a) at ({\a},{0});
    	\coordinate (b) at ({\b},{\b});
    	\coordinate (c) at ({0},{0});
    	\coordinate (aub) at ({\a},{\b-(\b-\a)*\al});
    	\coordinate (bub) at ({\b},{(\b-\a)*\al});
    	\coordinate (alb) at ({\a},{\a*\al});
    	\coordinate (blb) at ({\b},{\b*\al});
    	
    	\node[circle, fill=black, minimum size=3pt,inner sep=0pt, outer sep=0pt] at (a) {};
    	\node[circle, fill=black, minimum size=3pt,inner sep=0pt, outer sep=0pt] at (b) {};
    	\node[circle, fill=black, minimum size=3pt,inner sep=0pt, outer sep=0pt] at (c) {};
    	
    	\fill[fill=blue!90, opacity=0.3] (alb) -- (a) -- (b) -- (blb) -- cycle;
    	\draw[black,thick] (a) -- (c) -- (b);
    	\draw[-] (b) -- ($(b)+(0.3,0.3)$);
    	\draw[-] ({\a},-0.4) -- ({\a},0.4);
    	\draw[-] ({\b},-0.4) -- ({\b},1.9);
    	
    	\node[anchor=south east,align=center,scale=0.85] at ({\a},-0.50) {$l_x$};
    	\node[anchor=south west,align=center,scale=0.85] at ({\b},-0.50) {$u_x$};	
    	\node[anchor=south east,align=center,scale=0.85] at ({-0.1},0.70) {$y\leq \frac{u_x}{u_x-l_x} (x-l_x)$};
    	\node[anchor=north east,align=center,scale=0.85] at (-0.1,-0.1) {$y\geq \lambda x$};
    	
    	\node[anchor=south west,align=center,scale=0.85] at ($(b)+(0.2,-0.3)$) {$y=\max(0,x)$};
    
    \end{tikzpicture}
\vspace{-2mm}
	\caption{Linear bounds for $\text{ReLU}(x)$.}
    \label{fig:deeppoly_relu}
\end{wrapfigure}
\paragraph{Neural Network Verification} aims to decide whether the robustness properties defined above hold.
To this end, a wide range of methods has been proposed, many relying on bound propagation, i.e., determining a lower and upper bound for each neuron $l \leq x \leq u$, or in vector notation for the whole layer $\vl \leq \vx \leq \vu$.
Here, we discuss two ways of obtaining such bounds:
First, \emph{Interval Bound Propagation} \citep{GehrMDTCV18,MirmanGV18} where $\vl$ and $\vu$ are constants, bounding the reachable values of neurons. For details, we refer to \citet{GowalDSBQUAMK18}.
Second, \emph{Linear Bound Propagation} \citep{singh2019abstract,zhang2018crown}, where every layer's neurons $\vx_i$  are lower- and upper-bounded depending only on the previous layer's neurons:
\begin{equation}
	\label{eq:deeppoly}
	\mA_i^- \vx_{i-1} + \vc_i^- =: \vl_i \leq \vx_i, \quad \vx_i \leq \vu_i := \mA_i^+ \vx_{i-1} + \vc_i^+.
\end{equation}
Given these linear constraints, we can recursively substitute $\vx_{i-1}$ with its linear bounds in terms of $\vx_{i-2}$ until we have obtained bounds depending only on the input $\vx_0$. This allows us to compute concrete bounds $\vl$ and $\vu$ on any linear expression over network neurons.

For a linear layer $\vx_i = \mW_i \vx_{i-1} + \vb_i$ we simply have $\mA_i^\pm = \mW_i$ and $\vc_i^\pm = \vb_i$.
For a ReLU function $\vx_i = \text{ReLU}(\vx_{i-1})$, we first compute the input bounds $\vl \leq \vx_{i-1} \leq \vu$. If the ReLU is stably inactive, i.e. $u \leq 0$, we can replace it with the zero-function. If the ReLU is stably active, i.e. $l \geq 0$, we can replace it with the identity-function. In both cases, we can use the bounding for a linear layer. If the ReLU is unstable, i.e., $l < 0 < u$, we compute a convex relaxation with parameter $\lambda$ as illustrated in \cref{fig:deeppoly_relu}.
Using this backsubstitution approach, we can now lower bound the difference $y_t - y_i, \; \forall i \neq t$ to determine whether the target class logit $y_t$ is always greater than all other logits in the classification setting and similarly bound the elementwise output range in the regression setting.

\paragraph{Provable Training}
Special training is necessary to obtain networks that are provably robust. 
Considering the classification setting with a data distribution $(\vx, t) \sim \bc{D}$. Provable training now aims to choose the network parametrization $\bs{\theta}$ that minimizes the expected \emph{worst case} loss:
\begin{equation}
\label{eq:rob_opt}
\bs{\theta}_\text{rob} = \argmin_{\bs{\theta}} \mathbb{E}_{\bc{D}} \Big[ \max_{\vx' \in \bc{B}_p^{\epsilon_p}(\vx) }\bc{L}_\text{CE}(\vf_{\bs{\theta}}(\vx'),t) \Big]\quad \text{with} \quad \bc{L}_\text{CE}(\vy, t) = \ln\big(1 + \sum_{i \neq t} \exp(y_i-y_t)\big).
\end{equation}
The inner maximization problem is generally intractable, but can be upper bounded using bound propagation \citep{MirmanGV18,GowalDSBQUAMK18,zhang2019towards,mueller2023certified}.
\newpage

\section{Neural Ordinary Differential Equations} \label{sec:node}
In this section, we discuss the necessary background relating to \nodes \citep{chen2018neural}.

\paragraph{Neural Ordinary Differential Equations} are built around an initial value problem (IVP), defined by an input state $\vz(0) = \vz_{0}$ and a neural network $\vg_{\theta}$ defining the dynamics of an ordinary differential equation (ODE) $\nabla_t\vz(t) = \vg_{\theta}(\vz(t),t)$. We obtain its solution  $\vz(T)$ at time $T$ as
\begin{equation} \label{eq:NODE-sol}
    \vz(T) = \vz(0) + \int_{0}^{T} \vg_{\theta}(\vz(t),t) dt.
\end{equation}
Generally, $\vz_0$ can either be the raw input or come from an encoder neural network.
For both classification and regression tasks, we output $\vy = \vf_{\theta}(\vz(T_{\text{end}}))$ for an input $\vz_0$ and a predefined $T_{end}$, where $\vf$ is an additional decoder, usually a linear layer.

Time series forecasting is a special case of the regression setting where the input is a time-series $\vx^L_{ts} = \{(\vx_{j},t_{j})\}_{j=1}^L$, defined as a sequence of $L$ entries, each consisting of a data point $\vx_{j} \in \R^{d_\text{in}}$ and an observation time $t_{j}$. We aim to predict the value of the last observed data point $\vx_L$, using only the first $L' < L$ data points as input.
To this end, we employ the so-called latent-\ode architecture, where a recurrent encoder network reads the data sequence $\{(\vx_{j},t_{j})\}_{j=1}^{L'}$ and outputs the initial state $\vz_0$ for a decoder \node that is then integrated up to the desired time-step $T_{\text{end}} = t_L$ before its output $\vz_{t_L}$ is passed through a linear layer $\vf$. For further details, we refer to \cref{sec:latent_ode}.

\paragraph{\ode Solvers}
are employed to approximate \cref{eq:NODE-sol}, as analytical solutions often don't exist.
These solvers split the integration interval $[0,T]$ into sub-intervals, for which the integral is numerically approximated by evaluating $\vg_{\theta}$ 
at multiple points and taking their weighted average.
We let $\Gamma(\vz_0)$ denote the trajectory of an \ode solver, which we define as the sequence of tuples $(t,\ h)$ with time $t$ and step-size $h$.%

\ode solvers are characterized by their order $p$, indicating how quickly approximation errors diminish as the step size is reduced \citep{shampine2005error}. We distinguish between fixed ($h$ constant) \citep{euler1792institutiones,runge1895numerische} and adaptive solvers ($h$ varies; discussed below) \citep{dormand1980family,bogacki19893}. 
Note that for adaptive solvers, the trajectory depends on the exact input. %
\citet{huang2020adversarial} found that the supposedly inherent robustness of \nodes to adversarial attacks \citep{kang2021stable,yan2019robustness} is only observed for adaptive \ode solvers and may stem, partially or entirely, from gradient obfuscation \citep{athalye2018obfuscated} caused by the solver.

\paragraph{Adaptive \ode Solvers}
Adaptive step-size solvers \citep{dormand1980family,bogacki19893} use two methods of different order to compute the proposal solutions $\hat{\vz}^1(t+h)$ and $\hat{\vz}^2(t+h)$ and derive an error estimate $\delta = \|\frac{\hat{\vz}^1(t+h) - \hat{\vz}^2(t+h)}{\tau}\|_{1}$, normalized by the absolute error tolerance $\tau$.
This error estimate $\delta$ is then used to update the step size $h \leftarrow h \delta^{-\sfrac{1}{p}}$.
Next, we discuss the challenges this poses for robustness verification and how we tackle them.

\section{Controlled Adaptive \ode Solvers}\label{sec:controlled-adaptive}
\begin{wrapfigure}[12]{r}{0.22\textwidth}
\centering
\vspace{-7mm}
	\centering
	\includegraphics[width=1.0\linewidth]{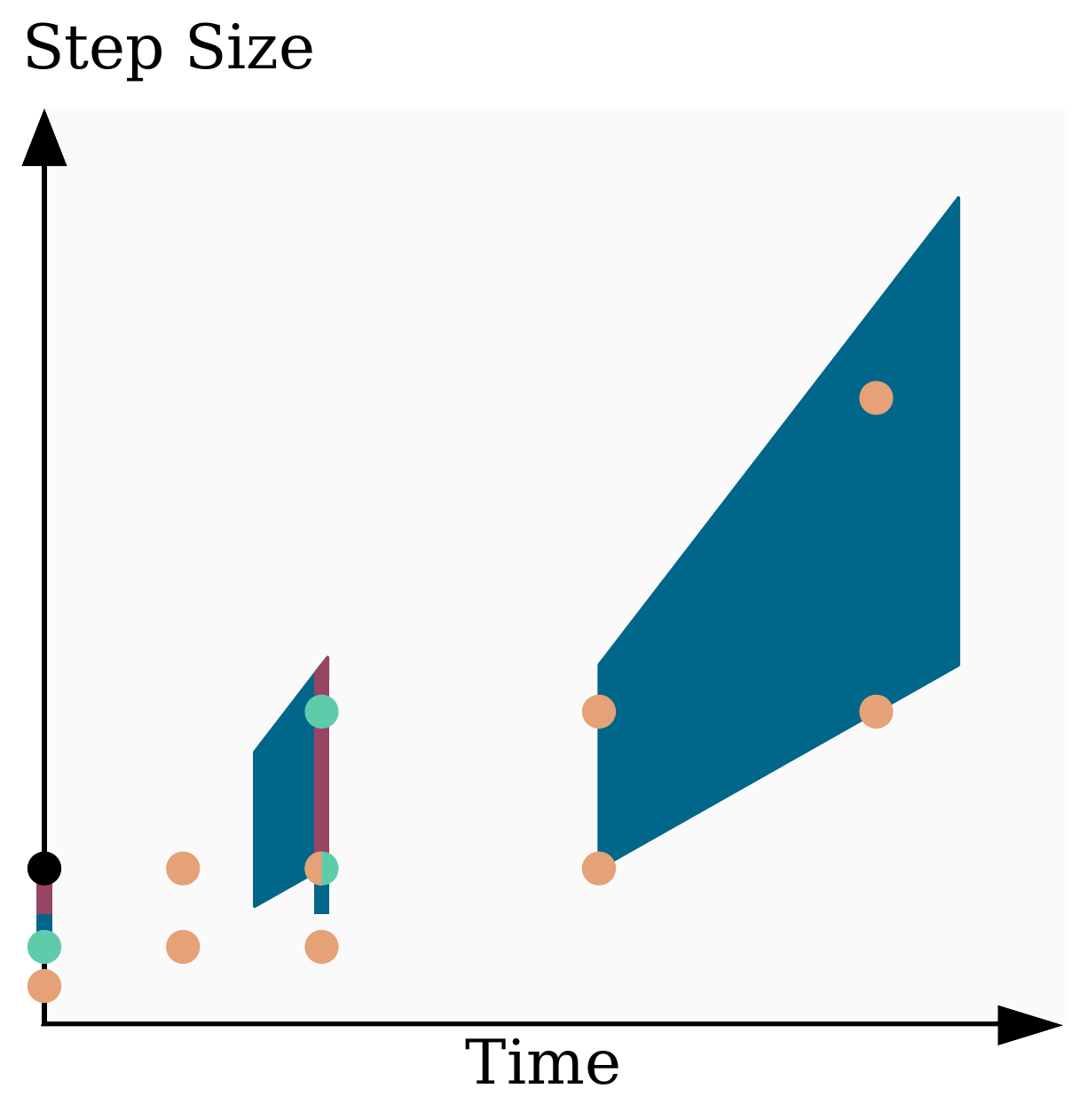}
	\vspace{-6.5mm}
	\caption[]{We compare \solver and adaptive solvers (AS) with respect to the reachable time/step-size tuples after one (\markerc{fteal}, \markerb{fred}) and two solver steps (\markerc{forange}, \markerb{fblue}).}
	\label{fig:CAS_time}
\end{wrapfigure}
Adaptive \ode solvers (AS) update their step-size $h$ continuously depending on the error estimate $\delta$. For continuous input regions, this generally yields infinitely many trajectories, making their abstraction intractable. We illustrate this in \cref{fig:CAS_time} (details in \cref{app:adaptive_solvers_comparison}), where the blue regions (\markerb{fblue}) mark all (time, step-size) tuples that are reachable after two steps.
To overcome this, we propose controlled adaptive solvers (\solver), which restrict step-sizes to a discrete set (\markerc{forange}), making them amenable to certification (\cref{sec:verification}). Next, we show how any adaptive \ode solver can be converted into a corresponding \solver solver.

\paragraph{Step-Size Update}
We modify the step-size update rule of any AS as 
\begin{equation*}%
    h  \leftarrow \begin{cases}
				h \cdot \alpha, \quad &\text{if } \delta \leq \tau_{\alpha} ,\\
				h, \quad &\text{if }  \tau_{\alpha} < \delta \leq 1, \\
				h / \alpha, \quad &\text{otherwise.}
\end{cases} \qquad \delta = \left\|\frac{\hat{\vz}^1(t+h) - \hat{\vz}^2(t+h)}{\tau}\right\|_{1},
\end{equation*}
with update factor $\alpha \in \mathbb{N}^{> 1}$, and the $\alpha$-induced decision threshold $\tau_{\alpha}= \alpha^{-p}$. 
Intuitively, we increase the step size by a factor $\alpha$ if we expect the normalized error after this increase to still be acceptable, i.e., $\delta \leq \alpha^{-p}$, we decrease the step size by a factor $\alpha$ and repeat the step if the error exceeds our tolerance, i.e., $\delta > 1$, and we keep the same step size otherwise. 
If the time $t+h$ after the next step would exceed the final time $T_\text{end}$, we clip the step size to $h \gets \min(h, T_{end} - t)$. Additionally, we enforce a minimum step-size. 
For more details, see \cref{app:adaptive_solvers}.

We contrast the update behaviors of \solver and AS solvers in \cref{fig:CAS_time}. We initialize both solvers with the same state (\markerc{black}) and after one step, the \solver solver can reach exactly three different states (\markerc{fteal}) while the adaptive solver can already reach continuous states (\markerb{fred}). After two steps this difference becomes even more clear with the \solver solver reaching only $9$ states (\markerc{forange}) while the adaptive solver can reach a large region of time/step-size combinations (\markerb{fblue}).

\paragraph{Initial Step-Size}
During training, the initial step size $h_0$ is computed based on the initial state and corresponding gradient. To avoid this dependence during inference, we always use its exponentially weighted average, computed during training \ccc{(details in \cref{app:adaptive_solvers})}. 

\begin{wrapfigure}[11]{r}{0.22\textwidth}
\centering
	\vspace{-3mm}
	\includegraphics[width=1.0\linewidth]{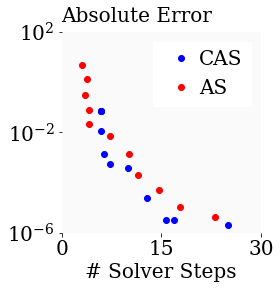} 
	\vspace{-6mm}
\caption[]{AS and \solver error over solver steps.}
\label{fig:CAS_error}
\end{wrapfigure}
\paragraph{Comparison to Adaptive Solvers}
\solver solvers can be seen as adaptive solvers with discretized step-sizes of the same order. Due to the exponentially spaced step-sizes, \solver can approximate any step-size chosen by an AS up to a factor of at most $\alpha$, with the \solver always choosing the smaller steps. Thus, \solver will need at most $\alpha$-times as many steps as an adaptive solver, assuming that the adaptive solver will never update the step size by more than $\alpha$ in one step.
Empirically, we confirm this on a conventional non-linear \ode, plotting mean absolute errors over the mean number of solver steps depending on the error threshold in \cref{fig:CAS_error}. 
There, we see that a dopri5-based \solver solver performs similarly to an unmodified dopri5 (AS). 
\ccc{For more details and additional comparisons between the solvers, we refer to \cref{app:adaptive_solvers_comparison} and \cref{app:additional_exp_CAS}.}

\section{Verification of Neural Ordinary Differential Equations} \label{sec:verification}
While the discrete step sizes of \solver, discussed in \cref{sec:controlled-adaptive}, yield a finite number of trajectories for any input region, there are still exponentially many in the integration time. Naively computing bounds for all of them independently is thus still intractable. 
To tackle this challenge, we introduce the analysis framework \tool, short for \tooll, which allows us to efficiently propagate bounds through the \ode solver using a graph representation of all trajectories. We discuss two instantiations, one using interval bounds, the other linear bounds.

Let us consider a \node with input $\bc{Z}$, either obtained from an encoder or directly from the data. We now define the trajectory graph $\bc{G}({\bc{Z}}) = (\bc{V}, \bc{E})$, representing all trajectories $\Gamma(\vz_0')$ for $\vz_0' \in \bc{Z}$.
The nodes $v \in \bc{V}$ represent solver states $(t, h)$ with time $t$ and step-size $h$ and aggregate interval bounds on the corresponding $\vz(t)$. The directed edges $e \in \bc{E}$ connect consecutive states in possible solver trajectories. This representation allows us to merge states $\vz(t)$ with identical time and step-size, regardless of the trajectory taken to reach them. 
This reduces the number of trajectories or rather solver steps we have to consider from exponential $\bc{O}(\exp(T_\text{end}))$ to quadratic $\bc{O}(T_\text{end}^2 \log^2(T_\text{end}))$ (given at most $\bc{O}(T_\text{end} \log(T_\text{end}))$ nodes in $\bc{V}$ \ccc{as derived in \cref{app:node_trafo_details}}), making the analysis tractable.

\paragraph{Verification with Interval Bounds}
We first note that each solver step only consists of computing the weighted sum of evaluations of the network $\vg$, allowing standard interval bound propagation to be used for its abstraction. We call this evaluation of a solver on a set of inputs an \emph{abstract solver step}.
Now, given an input $\bc{Z}$, we construct our trajectory graph $\bc{G}(\bc{Z})$ as follows: We do an abstract solver step, compute the interval bounds of the local error estimate $\delta_{(t, h)}$, and check which step size updates (increase, accept, or decrease) could be made according to the \solver. Depending on the looseness of the bounds, multiple updates might be chosen; we call this case trajectory splitting. For each possible update, we obtain a new state tuple $(t', h')$ and add the node $(t', h')$ to $\bc{V}$ and an edge from $(t, h)$ to $(t', h')$ to $\bc{E}$. If the node $(t', h')$ already existed, we update its state to contain the convex hull of the interval bounds.
We repeat this procedure until all trajectories have reached the termination node $(T_\text{end}, 0)$. This yields a complete trajectory graph and interval bounds for $\vz(T_\text{end})$.
If there are further layers after the \node, standard interval propagation can be employed to obtain the network output $\vy$.

\begin{figure*}
    \centering
    \begin{tikzpicture}[node/.style={circle,draw},scale=0.55,node distance={30mm},transform shape]
    \node[draw=black!60, fill=black!05, rectangle, rounded corners=2pt, minimum width=11.5cm, minimum height=5.2cm, anchor=north] (box) at (5.0, 0.0) {};
    \foreach \x in {0,1,2,3}
        \foreach \y in {1,2}
            \pgfmathsetmacro{\valid}{ifthenelse(\x+\y< 5,1,0)} 
            \def \t {$\frac{\x}{4}$}
            \ifnum \x= 0 \def \t {$0$} \fi
            \ifnum \x= 4 \def \t {$1$} \fi
            \ifnum \x= 2 \def \t{$\frac{1}{2}$} \fi
            \def \h {$\frac{\y}{4}$}
            \ifnum \y= 0 \def \h {$0$} \fi
            \ifnum \y= 4 \def \h {$1$} \fi
            \ifnum \y= 2 \def \h{$\frac{1}{2}$} \fi
            \ifnum \valid=1  \node[node] (\x\y) at (\x*2.5, \y*3 -7.35 ) {\t, \h} \fi; %
            
    \node[node] (end) at (10, 4.5 -7.35 ) {1, 0}; 
    \foreach \x in {0,1,2,3}
        \foreach \y  in {1,2}
            \pgfmathsetmacro{\valid}{ifthenelse(\x+\y< 5 ,1,0)}
            \ifnum \valid=1 
                \pgfmathsetmacro{\reject}{ifthenelse(\y==2,1,0)}
                \ifnum \reject= 1 \pgfmathsetmacro{\yn}{\y*0.5} \draw[-stealth] (\x\y.south) -- (\x1.north) \fi
            \fi;
    \foreach \x in {1,2,3}
        \foreach \y  in {1,2}
            \pgfmathsetmacro{\valid}{ifthenelse(\x+\y< 5 ,1,0)}
            \ifnum \valid=1 
                \pgfmathsetmacro{\accept}{ifthenelse(2*\y +\x<5,1,0)}
                \ifnum \accept = 1   \pgfmathtruncatemacro{\xn}{\y+\x)}  \draw[-stealth] (\x\y.east) -- (\xn\y.west) \fi
            \fi;
    \draw[-stealth] (01.east) -- (11.west); %
    \draw[-stealth] (02.north east) to [out=32,in=153]  (22.north west);
    \foreach \x in {0,1,2,3}
        \foreach \y  in {1,2}
            \pgfmathsetmacro{\valid}{ifthenelse(\x+\y< 5 ,1,0)}
            \ifnum \valid=1 
                \pgfmathsetmacro{\accept}{ifthenelse(\y +\x==4,1,0)}
                \ifnum \accept = 1   \pgfmathtruncatemacro{\xn}{\y+\x)}  \draw[-stealth] (\x\y.east) -- (end.west) \fi
            \fi;
    \foreach \x in {0,1,2,3}
        \foreach \y  in {1,2}
            \pgfmathsetmacro{\valid}{ifthenelse(\x+\y< 5 ,1,0)}
            \ifnum \valid=1 
                \pgfmathsetmacro{\increase}{ifthenelse(3*\y +\x<5,1,0)}
                \ifnum \increase = 1   \pgfmathtruncatemacro{\xn}{\y+\x)} \pgfmathtruncatemacro{\yn}{2*\y)}  \draw[-stealth] (\x\y.north east) -- (\xn\yn.south west) \fi
            \fi;
    \draw[-stealth] (12.east) -- (31.north west);
    \node[anchor=south east]() at (box.south east) {\textbf{(d)}};
    
    \begin{scope}[xshift = -12.4cm]
            \node[draw=black!60, fill=black!05, rectangle, rounded corners=2pt, minimum width=1.5cm, minimum height=5.2cm, anchor=north] (box) at (0., -0.0) {};
		\node[node, red!85!black!70] (01) at (0*2.0, 2*3 -7.35 ) {0, $\frac{1}{2}$};
		\node[anchor=south east]() at (box.south east) {\textbf{(a)}};
	\end{scope}
    \begin{scope}[xshift = -10.4cm]
        \node[draw=black!60, fill=black!05, rectangle, rounded corners=2pt, minimum width=3.95cm, minimum height=5.2cm, anchor=north] (box) at (1.2, -0.0) {};
	    \node[node, blue!85!black!80] (02) at (0*2.0, 2*3 -7.35 ) {0, $\frac{1}{2}$};
	    \node[node, red!85!black!70] (01) at (0*2.0, 1*3 -7.35 ) {0, $\frac{1}{4}$};
	    \node[node, red!85!black!70] (22) at (1.2*2.0, 2*3 -7.35 ) {$\frac{1}{2}$, $\frac{1}{2}$};
	    
	    \draw[-stealth, red!85!black!70] (02.south) -- (01.north);
	    \draw[-stealth, red!85!black!70] (02.north east) to [out=45,in=145]  (22.north west);
	    \node[anchor=south east]() at (box.south east) {\textbf{(b)}};
    \end{scope}
    \begin{scope}[xshift = -6cm]
        \node[draw=black!60, fill=black!05, rectangle, rounded corners=2pt, minimum width=5.5cm, minimum height=5.2cm, anchor=north] (box) at (2.0, 0.0) {};
	    \node[node] (02) at (0*2.0, 2*3 -7.35 ) {0, $\frac{1}{2}$};
	    \node[node, blue!85!black!80] (01) at (0*2.0, 1*3 -7.35 ) {0, $\frac{1}{4}$};
	    \node[node] (22) at (2*2.0, 2*3 -7.35 ) {$\frac{1}{2}$, $\frac{1}{2}$};
	    \node[node, red!85!black!70] (12) at (1*2.0, 2*3 -7.35 ) {$\frac{1}{4}$, $\frac{1}{2}$};
	    \node[node, red!85!black!70] (11) at (1*2.0, 1*3 -7.35 ) {$\frac{1}{4}$, $\frac{1}{4}$};
	    
	    \draw[-stealth] (02.south) -- (01.north);
	    \draw[-stealth] (02.north east) to [out=45,in=145]  (22.north west);
	    \draw[-stealth, red!85!black!70] (01.east) -- (11.west);
	    \draw[-stealth, red!85!black!70] (01.north east) to (12.south west);
	    \node[anchor=south east]() at (box.south east) {\textbf{(c)}};
    \end{scope}
    \end{tikzpicture}
     %
    \vspace{-5mm}
    \caption{ An example trajectory graph $\mathcal{G}(\bc{Z})$ construction for a controlled adaptive ODE solver with $h_0 = \frac{1}{2}$, $\alpha = 2$ and $T_{end} = 1$. Note how trajectory splitting occurs in all vertices except the last two states.} %
    \label{fig:trajectory-graph}
    \vspace{-4mm}
\end{figure*}
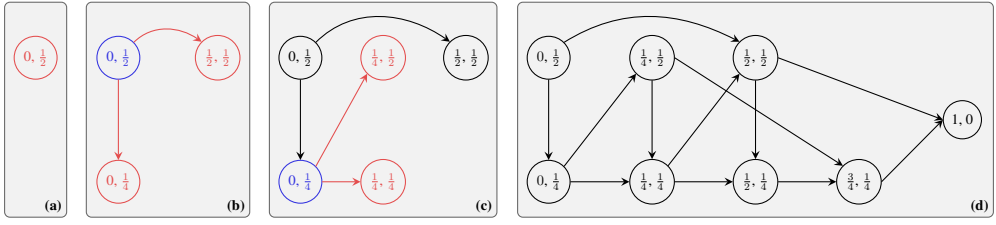

We illustrate this construction process in \cref{fig:trajectory-graph}, where we highlight newly added edges and nodes in red and the processed node in blue: We initialize the graph with the node $(0, h_0)$, in our case $h_0 = \sfrac{1}{2}$ (see \cref{fig:trajectory-graph}(a)). We now do an abstract solver step for this node and find that $\delta > \tau_\alpha$. Thus, we either accept the step, yielding the next node $(\sfrac{1}{2}, \sfrac{1}{2})$, or we reject the step and decrease the step-size by $\alpha=2$, yielding the node $(0, \sfrac{1}{4})$, both are connected to the current node (see \cref{fig:trajectory-graph}~(b)). We now choose among the nodes without outgoing edges the one with the smallest current time $t$ and largest step-size $h$ (in that order), $(0,\sfrac{1}{4})$ in our case, and do another abstract solver step, yielding $\delta < 1$. We thus either accept the step, yielding the node $(\sfrac{1}{4}, \sfrac{1}{4})$, or additionally increase the step-size, yielding the node $(\sfrac{1}{4},\sfrac{1}{2})$ (see \cref{fig:trajectory-graph} (c)). We proceed this way until the only node without outgoing edges is the termination node $(T_\text{end},0)$ with $T_\text{end} = 1$ in our case (see \cref{fig:trajectory-graph} (d)).

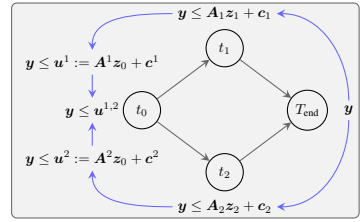
\begin{wrapfigure}[14]{r}{0.35\textwidth}
\vspace{-4mm}
\centering
\begin{tikzpicture}[node/.style={circle,draw},scale=0.55,node distance={30mm},transform shape,minimum width=0.9cm]
        \node[draw=black!60, fill=black!05, rectangle, rounded corners=2pt, minimum width=8.4cm, minimum height=5.2cm, anchor=north] (box) at (0.55, 2.6) {};
	    \node[node] (A) at (-0.5, 0) {$t_0$};
	    \node[node] (B) at (1.5, 1.5) {$t_1$};
	    \node[node] (C) at (1.5, -1.5) {$t_2$};
	    \node[node] (D) at (3.5, 0) {$T_{\text{end}}$};

	    \node[right=0.5mm of D ] (D_bound)  {$\vy$};
	    \node[above=1mm of B ] (B_bound)  {$\vy \leq \mA_1 \vz_1 + \vc_{1}$ };
	    \node[below=1mm of C ] (C_bound)  {$\vy \leq \mA_2 \vz_2 + \vc_{2}$ };

	    \node[above=4mm of A, xshift=-12mm ] (AB_bound)  {$\vy \leq \vu^1 := \mA^1 \vz_0 + \vc^1$ };
	    \node[below=4mm of A, xshift=-12mm ] (AC_bound)  {$\vy \leq \vu^2 := \mA^2 \vz_0 + \vc^2$ };
	    \node (A_bound) at (AB_bound |- A)  {$\vy \leq \vu^{1,2}$ };

	    \draw[-stealth, black!60] (A) -- (B);
	    \draw[-stealth, black!60] (A) -- (C);
	    \draw[-stealth, black!60] (B) -- (D);
	    \draw[-stealth, black!60] (C) -- (D);

	    \draw[-stealth, blue!60] (D_bound) to [out=90, in=0] (B_bound.east);
	    \draw[-stealth, blue!60] (D_bound) to [out=270, in=0] (C_bound.east);

	    \draw[-stealth, blue!60] (B_bound) to [out=180, in=90] (AB_bound.north);
	    \draw[-stealth, blue!60] (C_bound) to [out=180, in=270] (AC_bound.south);

	    \draw[-stealth, blue!60] (AB_bound.south) -- (A_bound.north);
	    \draw[-stealth, blue!60] (AC_bound.north) -- (A_bound.south);
\end{tikzpicture}
\vspace{0mm}
\caption{Example upper bounds for $\vy = \vz(T_\text{end})$ via \tool. (Lower bounds analogous.) Blue arrows show the backward substitution resulting in \lcap at $t_0$.} \label{fig:multiple_constraints}
\end{wrapfigure}
\paragraph{Verification with Linear Bounds}
To compute more precise linear bounds on $\vz(T_{\text{end}})$, we first construct the trajectory graph $\bc{G}(\bc{Z})$ as discussed above, using either interval bounds or the linear bounding procedure described below, retaining concrete element-wise upper and lower bounds at every state.
We can now derive linear bounds on $\vz(T_{\text{end}})$ in terms of the \node input $\vz_0$ by recursively substituting bounds from intermediate computation steps.
Starting with the bounds for $(T_{\text{end}}, 0)$, we backsubstitute them along every incoming edge, yielding a set of bounds in every preceding node. We recursively repeat this procedure until we arrive at the input node. 
We illustrate this in \cref{fig:multiple_constraints}, where we, starting at $T_\text{end}$, backsubstitute $\vy$ to $t_1$ and $t_2$, obtaining bounds in terms of $\vz_1$ and $\vz_2$.
In contrast to the standard \deeppoly backward substitution procedure, a node in $\bc{G}(\bc{Z})$ can have multiple successors which reach the final node via different trajectories. We can thus obtain several sets of linear constraints bounding the same expression with respect to the same state, which we need to merge in a sound manner without losing too much precision.
We call this the linear constraint aggregation problem (\lcap) and observe that it arises in \cref{fig:multiple_constraints} after an additional backsubstitution step to $t_0$ yields two bounds, $\vu^1$ and $\vu^2$, on $\vy$ both in terms of $\vz_0$.

\begin{wrapfigure}[11]{r}{0.42\textwidth}
    \centering
    \vspace{-7mm}
    \begin{tikzpicture}[node/.style={circle,draw},scale=0.65,node distance={30mm},transform shape]

    \begin{scope}[xshift = -12.5cm]
            \node[draw=black!0, fill=black!0, rectangle, rounded corners=2pt, minimum width=4cm, minimum height=4cm, anchor=south west] (box) at (-0.8, -1.0) {};
            
        \draw[-{Stealth[length=2mm, width=1.4mm]}] (-0.4, -0.2) -- (2.7, -0.2) node[right,scale=1] {$z_0$};
    	\draw[-{Stealth[length=2mm, width=1.4mm]}] (-0.3, -0.3) -- (-0.3, 2.2) node[above,scale=1] {$y$};
        \def\scaley{1.5}
    	\def\a{0}
    	\def\b{1.5}
    	\def\al{0.15}
    	\coordinate (r1_l) at ({0},{1*\scaley});
    	\coordinate (r1_u) at ({2},{(1 + 2 * (-0.4))*\scaley });
    	\draw[-,blue] (r1_l) -- (r1_u);
    	
    	\coordinate (r2_l) at ({0},{0});
    	\coordinate (r2_u) at ({2},{(2 * (0.6))*\scaley });
    	\draw[-,orange] (r2_l) -- (r2_u);
    	
    	\coordinate (r3_l) at ({0},{0.5*\scaley});
    	\coordinate (r3_u) at ({2},{(0.5 + 2 * (0.25))*\scaley });
    	\draw[-,violet] (r3_l) -- (r3_u);
    	
    	\draw[-] (0,-0.3) -- (0,-0.1);
    	
    	\draw[-] (2,-0.3) -- (2,-0.1);

  		\node[anchor=north]() at ($(box.south)+(-0.150,0.7)$) {\textbf{(a)}};   
	\end{scope}
    \begin{scope}[xshift = -8.5cm]
        \node[draw=black!0, fill=black!0, rectangle, rounded corners=2pt, minimum width=4cm, minimum height=4cm, anchor=south west] (box) at (-0.8, -1.0) {};
        
        \draw[-{Stealth[length=2mm, width=1.4mm]}] (-0.4, -0.2) -- (2.7, -0.2) node[right,scale=1] {$z_0$};
    	\draw[-{Stealth[length=2mm, width=1.4mm]}] (-0.3, -0.3) -- (-0.3, 2.2) node[above,scale=1] {$y$};
        \def\scaley{1.5}
    	\def\a{0}
    	\def\b{1.5}
    	\def\al{0.15}
    	\coordinate (r1_l) at ({0},{1*\scaley});
    	\coordinate (r1_u) at ({2},{(1 + 2 * (-0.4))*\scaley });
    	\draw[-,densely dashed,blue] (r1_l) -- (r1_u);
    	
    	\coordinate (r2_l) at ({0},{0});
    	\coordinate (r2_u) at ({2},{(2 * (0.6))*\scaley });
    	\draw[-,densely dashed,orange] (r2_l) -- (r2_u);
    	
    	\coordinate (r3_l) at ({0},{0.5*\scaley});
    	\coordinate (r3_u) at ({2},{(0.5 + 2 * (0.25))*\scaley });
    	\draw[-,violet, dotted] (r3_l) -- (r3_u);
    	
    	\coordinate (r4_l) at ({0},{1*\scaley});
    	\coordinate (r4_u) at ({2},{(1 + 2 * (0.1))*\scaley });
    	\draw[-,green] (r4_l) -- (r4_u);
    	
    	\draw[-] (0,-0.3) -- (0,-0.1);
    	
    	\draw[-] (2,-0.3) -- (2,-0.1);

  		\node[anchor=north]() at ($(box.south)+(-0.15,0.7)$) {\textbf{(b)}};   

    \end{scope}

    \begin{scope}[xshift = -12.0cm, yshift=-1.0cm]
	\def \shifty{0.7}

	\node[align=center,scale=1] at (0.8, 0) {\large $\vu^{1}$};
    	\draw[-,orange] (0,0) -- (0.3,0);

	\node[align=center,scale=1] at (0.8 + 1.5, 0) {\large $\vu^{2}$};
    	\draw[-,blue] (0 + 1.5,0) -- (0.3 + 1.5,0);

 	\node[align=center,scale=1] at (0.8 + 3.0, 0) {\large $\vu^{3}$};
     	\draw[-,violet] (0 + 3,0) -- (0.3 + 3,0);

	\node[align=center,scale=1] at (0.8 + 4.5, 0) {\large $\vu^{1,2}$};
    	\draw[-,green] (0 + 4.5,0) -- (0.3 + 4.5, 0);

    \end{scope}

    \end{tikzpicture} %
    \vspace{-9mm}
    \caption{Visualization of the \lcap with $m=3$, shown in \textbf{(a)}. In \textbf{(b)} the constraints $\vu^1, \vu^2$ (dashed) are over-approximated by $\vu^{1,2}$ via \lcapm, which also bounds $\vu^3$ (dotted).} \label{fig:aggregation}
\end{wrapfigure}
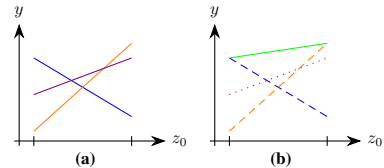
\paragraph{Linear Constraint Aggregation Problem}\label{sec:DPCA}
The \lcap requires us to soundly merge a set of different linear constraints bounding the same variable.
As an example, we consider a variable $y = \vz(T)$ for which we have $m$ upper bounds $\{\vu^j\}_{j=1}^m$ linear in $\vz_0$, which in turn can take values in $\bc{Z}$. In this case, we want to obtain a single linear upper bound $y \leq \va \vz_0 + c$ that minimizes the volume between the constraint and the $y=0$ plane over $\vz_0 \in \bc{Z}$, while soundly over-approximating all constraints.
More formally, we want to solve:
\begin{equation} \label{eq:LCAP}
\argmin_{\va, c} \int_{\bc{Z}} \va \vz_0 + c \,\mathrm{d}\vz_0, \quad s.t. \;\; \va \vz_0 + c \geq \max_j \va^j \vz_0 + c^j, \quad \forall \vz_0 \in \bc{Z} .
\end{equation}
While this can be cast as a linear program by enumerating all exponentially many corners of $\bc{Z}$, this becomes intractable even in modest dimensions.
To overcome this challenge, we propose \lcapml (\lcapm), translating the $\max \{\vu^j\}_{j=1}^m$ into a composition of ReLUs, which can be handled using the efficient \deeppoly primitive proposed by \citet{singh2019abstract}.
For a pair of constraints $\vu_i^1, \vu_i^2$ we can rewrite their maximum as 
\begin{equation}
	\max_{j \in {1, 2}} \vu_i^j = \vu_i^1 + \max(0, \vu_i^2 - \vu_i^1) = \vu_i^1 + \relu(\vu_i^2 - \vu_i^1).
\end{equation}
In the case of $m$ constraints, this rewrite can be applied multiple times.
We note that lower bounds can be merged analogously and visualize \lcapm for a $1$-d problem in \cref{fig:aggregation}. There,  the first iteration already yields the constraint $\vu^{1,2}$, dominating the remaining $\vu^{3}$.

\paragraph{Training}

In order to train \nodes amenable to verification we utilize the box bounds discussed above and sample $\kappa$ trajectories form $\bc{G}(\bc{Z})$.
For more details, please see \cref{app:node_trafo_details}.

\ccc{\paragraph{Bound Calculation}
During the computation of the bounds, \tool combines verification with interval and linear bounds by using the tighter bound of either approach (more details in \cref{app:verification_methods}).}

\section{Experimental Evaluation} \label{sec:experiments}

\paragraph{Experimental Setup}
We implement \tool in PyTorch\footnote{We release our code at \url{https://github.com/eth-sri/GAINS}} \citep{PaszkeGMLBCKLGA19} and evaluate all benchmarks using single NVIDIA RTX 2080Ti. We conduct experiments on \mnist \citep{LeCunBBH98}, \fmnist \citep{XiaoFashion17}, and \physio \citep{silva2012predicting}. For image classification, we use an architecture consisting of two convolutional and one NODE layer (see \cref{tab:architecture-classification} in \cref{app:classification-details} for more details). For time-series forecasting, we use a latent ODE (see \cref{tab:architecture-forecasting} in \cref{app:forecasting-model} for more details). We provide detailed hyperparameter choices in \cref{app:classification-details,app:time-series-forecasting-details}.

\begin{table}[]
    \centering	
    \begin{adjustbox}{width=\columnwidth,center}
    \begin{threeparttable}
    	\renewcommand{\arraystretch}{0.98}
    \caption{Means and standard deviations of the standard (Std.), adversarial (Adv.), and certified (Cert.) accuracy obtained with \tool depending on the training method and evaluated on the first 1000 test set samples.\vspace{-2mm}}
    \begin{tabular}{llccccccccc}
		\toprule
       	 \multirow{2.5}*{Dataset} & \multirow{2.5}*{Training Method} & \multirow{2.5}*{$\epsilon_t$} &  & \multirow{2.5}*{Std. [\%]} & \multicolumn{2}{c}{$\epsilon = 0.10$} & \multicolumn{2}{c}{$\epsilon = 0.15$} & \multicolumn{2}{c}{$\epsilon = 0.20$}\\
       	 \cmidrule(rl){6-7} \cmidrule(rl){8-9} \cmidrule(rl){10-11}
       	 &&&&& Adv. [\%] & Cert. [\%] & Adv. [\%] & Cert. [\%]& Adv. [\%] & Cert. [\%]\\
         \midrule
         \multirow{5}*{\mnist}& Standard & & & 98.8$^{\pm 0.4}$&23.2$^{\pm 3.5}$&0.0$^{\pm 0.0}$&2.5$^{\pm 1.6}$&0.0$^{\pm 0.0}$&0.3$^{\pm 0.2}$&0.0$^{\pm 0.0}$ \\
         \cmidrule(rl){2-3}
                              & Adv. & 0.11 & &  \textbf{99.2}$^{\pm 0.1}$& \textbf{95.4}$^{\pm 0.4}$& 0.0$^{\pm 0.0}$& \textbf{88.3}$^{\pm 0.6}$& 0.0$^{\pm 0.0}$& 59.4$^{\pm 3.2}$&0.0$^{\pm 0.0}$ \\
         \cmidrule(rl){2-3}
                              & \multirow{2}*{\toolt} & 0.11    & & 95.5$^{\pm 0.1}$&91.5$^{\pm 0.6}$&\textbf{89.0}$^{\pm 1.1}$&84.0$^{\pm 2.7}$&47.2$^{\pm 7.9}$&21.4$^{\pm 1.8}$&0.2$^{\pm 0.2}$ \\
                              & & 0.22    & &91.8$^{\pm 1.3}$&88.5$^{\pm 1.8}$&86.8$^{\pm 2.0}$&86.8$^{\pm 2.1}$&\textbf{83.7}$^{\pm 2.3}$&\textbf{84.5}$^{\pm 3.2}$&\textbf{79.7}$^{\pm 3.4}$ \\
        \cmidrule(rl){1-3}
        \multirow{5}*{\fmnist}& Standard &&& \textbf{88.6}$^{\pm1.2}$&0.1$^{\pm 0.1}$&0.0$^{\pm 0.0}$&0.0$^{\pm 0.0}$&0.0$^{\pm 0.0}$ & & \\
		\cmidrule(rl){2-3}
								& Adv. & 0.11 & & 80.9$^{\pm 0.7}$&\textbf{70.2}$^{\pm 0.5}$&0.0$^{\pm 0.0}$ &47.1$^{\pm 3.7}$&0.0$^{\pm 0.0}$& &  \\
		\cmidrule(rl){2-3}
								& \multirow{2}*{\toolt} & 0.11    &  & 75.1$^{\pm 1.2}$&65.7$^{\pm 1.0}$&\textbf{62.5}$^{\pm 1.1}$&21.1$^{\pm 5.9}$&13.3$^{\pm 3.1}$ & & \\
								& & 0.16    &  & 71.5$^{\pm 1.7}$&64.0$^{\pm 2.7}$&61.3$^{\pm 2.7}$&\textbf{60.1}$^{\pm 3.5}$&\textbf{55.0}$^{\pm 4.3}$& &  \\
         \bottomrule
    \end{tabular}
    \label{tab:results_cls}
    \end{threeparttable}
	\end{adjustbox}
\vspace{-4mm}
\end{table}

\subsection{Classification} \label{sec:classif-exp}
We train NODE based networks with standard, adversarial, and provable training ($\epsilon_t \in \{0.11, 0.22\}$) and certify robustness to $\ell_\infty$-norm bounded perturbations of radius $\epsilon$ as defined in \cref{eq:adv_robustness_cls}. We report means and standard deviations across three runs at different perturbation levels ($\epsilon \in \{0.1, 0.15, 0.2\}$) depending on the training method in \cref{tab:results_cls}.
Both for \mnist and \fmnist, adversarial accuracies are low ($0.0\%$ to $23.2\%$) for standard trained NODEs, agreeing well with recent observations showing vulnerabilities to strong attacks \citep{huang2020adversarial}. While adversarial training can significantly improve robustness even against these stronger attacks, we can not certify any robustness.
Using provable training with \tool significantly improves certifiable accuracy (to up to $89\%$ depending on the setting) while reducing standard accuracy only moderately. This trade-off becomes more pronounced as we consider increasing perturbation magnitudes for training and certification.

\begin{table}[h]
    \centering	
    \begin{adjustbox}{width=\columnwidth,center}
    \begin{threeparttable}
    \caption{Comparison of the mean absolute errors for the unperturbed samples (Std. MAE), and the adversarial (Adv.), and certifiable (Cert.) $\nu$-$\delta$-robustness with $\nu=0.1$ and $\delta=0.01$  obtained using different provable training methods on the full \physio test set.\vspace{-1mm}}
    \begin{tabular}{llccccccccc}
		\toprule
       	 \multirow{2.5}*{Setting} & \multirow{2.5}*{Training Method} & \multirow{2.5}*{$\epsilon_t$} &  & \multirow{2.5}*{Std. MAE  [$\times 10^{-2}$]} & \multicolumn{2}{c}{$\epsilon = 0.05$} & \multicolumn{2}{c}{$\epsilon = 0.10$} & \multicolumn{2}{c}{$\epsilon = 0.20$}\\
       	 \cmidrule(rl){6-7} \cmidrule(rl){8-9} \cmidrule(rl){10-11}
       	 &&&&& Adv. [\%] & Cert. [\%] & Adv. [\%] & Cert. [\%]& Adv. [\%] & Cert. [\%]\\
         \midrule
         \multirow{3.5}*{6h}& Standard &  & & \textbf{47.4}$^{\pm 0.3}$  & 54.3$^{\pm 3.8}$ & 0.0$^{\pm 0.0}$&13.7$^{\pm 2.9}$&0.0$^{\pm 0.0}$ &2.3$^{\pm 1.2}$ &0.0$^{\pm 0.0}$ \\
         \cmidrule(rl){2-3}
                              & \multirow{2}*{\toolt} & 0.1 && 51.1$^{\pm 2.0}$  & 97.7$^{\pm 0.7}$&  93.0$^{\pm 2.7}$  &77.0$^{\pm 7.3}$  & 60.4$^{\pm 10.9}$ & 42.0$^{\pm 11.0}$ &  24.2$^{\pm 7.7}$ \\
                              &                       & 0.2 && 57.6$^{\pm 2.5}$  & \textbf{100.0}$^{\pm 0.0}$&  \textbf{99.8}$^{\pm 0.1}$  &\textbf{96.4}$^{\pm 2.1}$ & \textbf{93.1}$^{\pm 4.5}$ & \textbf{80.1}$^{\pm 11.7}$ &  \textbf{70.5}$^{\pm 18.9}$\\
        \cmidrule(rl){1-3}
         \multirow{3.5}*{12h}& Standard & & & \textbf{49.9}$^{\pm 0.2}$& 65.2$^{\pm 2.0}$&0.0$^{\pm 0.0}$ &16.6$^{\pm 2.3}$&0.0$^{\pm 0.0}$& 2.0$^{\pm 0.4}$&0.0$^{\pm 0.0}$ \\
		\cmidrule(rl){2-3}
							& \multirow{2}*{\toolt} & 0.1 && 50.9$^{\pm 0.4}$  & 98.0$^{\pm 0.2}$&  94.5$^{\pm 0.7}$  &74.3$^{\pm 3.5}$ & 55.8$^{\pm 1.5}$ & 28.9$^{\pm 3.6}$&  17.2$^{\pm 0.1}$ \\
							&                       & 0.2 && 52.9$^{\pm 0.1}$  & \textbf{99.1}$^{\pm 0.1}$&  \textbf{98.3}$^{\pm 0.2}$  & \textbf{87.8}$^{\pm 0.8}$& \textbf{80.3}$^{\pm 0.8}$ &\textbf{52.3}$^{\pm 0.8}$ &  \textbf{38.5}$^{\pm 1.7}$ \\
		\cmidrule(rl){1-3}
    	\multirow{3.5}*{24h}& Standard & & &\textbf{51.2}$^{\pm 0.3}$& 69.7$^{\pm 1.9}$ &0.0$^{\pm 0.0}$ &23.6$^{\pm 2.8}$&0.0$^{\pm 0.0}$& 3.7$^{\pm 1.0}$&0.0$^{\pm 0.0}$ \\
		\cmidrule(rl){2-3}
							& \multirow{2}*{\toolt} & 0.1 && 51.5$^{\pm 0.1}$  &97.9$^{\pm 0.2}$&  96.2$^{\pm 0.4}$  & 78.3$^{\pm 2.3}$ & 68.0$^{\pm 1.6}$ & 32.6$^{\pm 0.6}$ &  22.7$^{\pm 1.0}$ \\
							&                       & 0.2 && 53.7$^{\pm 0.7}$  &\textbf{99.7}$^{\pm 0.1}$ &  \textbf{99.1}$^{\pm 0.3}$  & \textbf{92.3}$^{\pm 1.7}$ & \textbf{89.4}$^{\pm 2.4}$ & \textbf{59.8}$^{\pm 7.7}$&  \textbf{50.5}$^{\pm 5.1}$ \\
         \bottomrule
    \end{tabular}
    \label{tab:results_time}
    \end{threeparttable}
	\end{adjustbox}
	\vspace{-3mm}
\end{table}
\subsection{Time-Series Forecasting} \label{sec:forecasting-exp}
For time-series forecasting, we consider the \physio \citep{silva2012predicting} dataset, containing $8\,000$ time-series of up to $48$ hours of $35$ irregularly sampled features. We rescale most features to mean $\mu=0$ and standard deviation $\sigma=1$ (before applying perturbations) and refer to \cref{app:time-series-forecasting-details} for more details.
We consider three settings, where we predict the last measurement $L$, without having access to the preceding $6$, $12$, or $24$ hours of data. 
In \cref{tab:results_time}, we report the mean absolute prediction error (MAE) for the unperturbed samples and $\nu$-$\delta$-robustness (see \cref{eq:adv_robustness_reg}) for relative and absolute error tolerances of $\nu = 0.1$ and $\delta=0.01$, respectively, at perturbation magnitudes $\epsilon = \{0.05, 0.1, 0.2\}$.
We observe only a minimal drop in standard precision, when certifiably training with \tool at moderate perturbation magnitudes ($\epsilon_t=0.1$) while increasing both adversarial and certified accuracies substantially. 
Interestingly, the drop in standard precision is the biggest for the $6h$ setting, despite having the shortest forecast horizon among all settings.
We hypothesize that this is due to the larger number of input points and thus abstracted embedding steps leading to increased approximation errors. 
Further, while we can again not verify any robustness for standard trained \nodes, they exhibit non-vacuous empirical robustness. However, without guarantees it remains unclear whether this is due to adversarial examples being harder to find or \nodes being inherently more robust. %
Across settings, we observe that training with larger perturbation magnitudes leads to slightly worse performance on unperturbed data, but significantly improves robustness.

\subsection{Ablation} \label{sec:trajectory-attacks}

\begin{wraptable}[10]{r}{0.45\textwidth}
	\vspace{-4.5mm}
	\caption{\footnotesize Mean and standard deviation of the attack success [\%] on the first 1000 samples of the \mnist test set.}
	\vspace{-3mm}
	\centering
		\resizebox{1.00\linewidth}{!}{
	\begin{tabular}{ccccc}
		\toprule
		Training                  & $\epsilon_t$  & \multicolumn{3}{c}{Attack Success [\%]}                      \\
		\cmidrule(rl){3-5}
		&                          & $\epsilon = 0.1$   & $\epsilon = 0.15$  & $\epsilon = 0.2$   \\
		\midrule
		Standard                  &                       & 98.9$^{\pm 0.3}$&100.0$^{\pm 0.1}$&100.0$^{\pm 0.0}$                 \\
		\cmidrule(rl){1-2}
		Adversarial               & 0.11                   & 99.3$^{\pm 0.1}$   & 100.0$^{\pm 0.0}$             & 100.0$^{\pm 0.0}$                \\
		\cmidrule(rl){1-2}
		\multirow{3}{*}{\tool} & 0.11                   & 73.4$^{\pm 3.5}$&86.3$^{\pm 3.5}$&95.5$^{\pm 1.8}$ \\
		& 0.22                   & 65.2$^{\pm 7.5}$&75.3$^{\pm 6.2}$&82.2$^{\pm 5.0}$ \\
		\bottomrule
	\end{tabular}
	\label{tab:traj-attack}
}
\end{wraptable}
\paragraph{Trajectory Sensitivity}
We investigate whether the solver trajectory, i.e., the chosen step-sizes, of \solver solvers are susceptible to adversarial perturbations by conducting an adversarial attack aiming directly to change the trajectory $\Gamma(\vz_0)$ (see \cref{app:trajectory-attack} for more details).
In \cref{tab:traj-attack}, we report the success rate of this attack for \mnist, showing that even at moderate perturbation magnitudes ($\epsilon=0.1$) attacks are (almost) always successful if models are trained using standard or adversarial training. While training with \tool reduces this susceptibility notably, it remains significant. This highlights the need to consider the effect of a chosen solver on robustness, motivating both the use of \solver solvers and the trajectory graph-based approach of \tool.

\begin{wrapfigure}[12]{r}{0.60\textwidth}
	\centering
	\begin{subfigure}{0.45\linewidth}
		\centering
		\includegraphics[width = \textwidth]{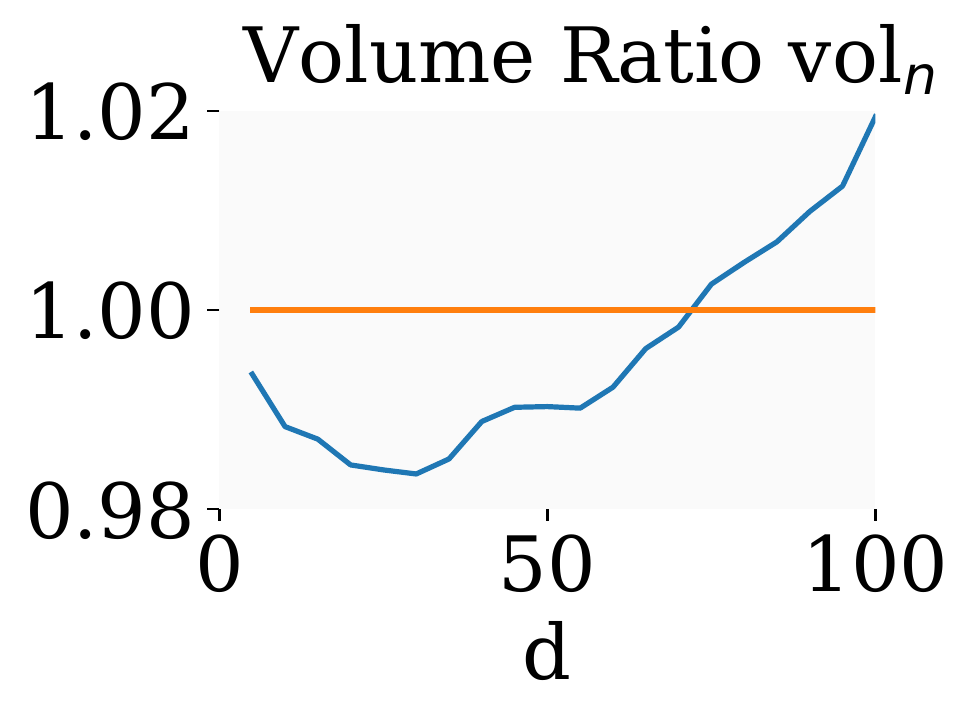}  
	\end{subfigure}
	\begin{subfigure}{0.45\linewidth}
		\vspace{-1mm}
		\centering
		\includegraphics[width = \textwidth]{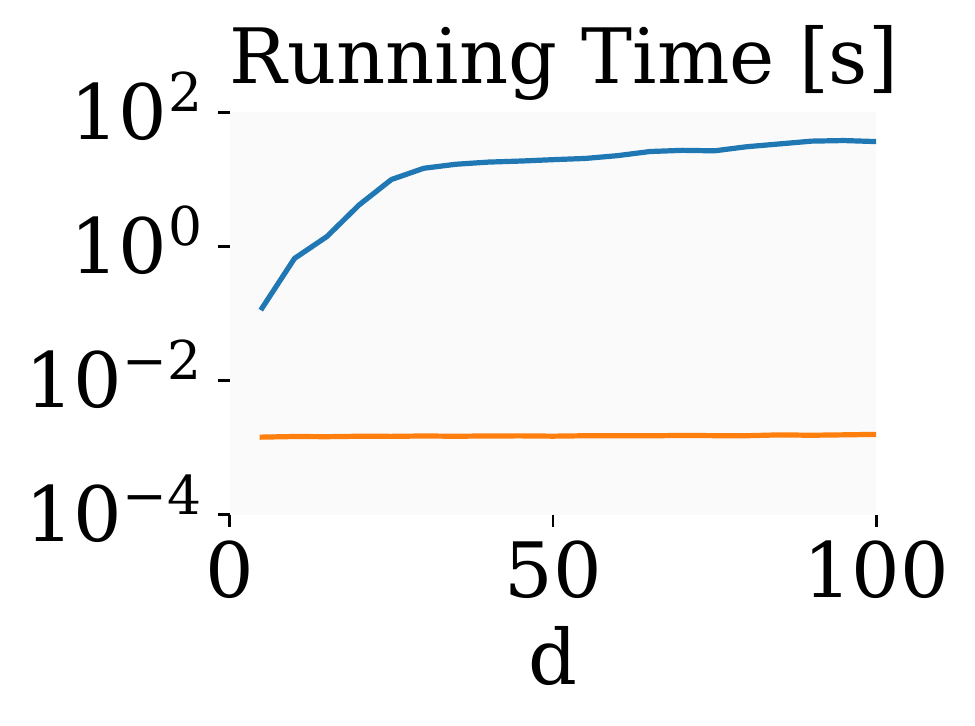}  
	\end{subfigure}
	\begin{subfigure}{0.45\linewidth}
		\centering
		\vspace{-8mm}
		\includegraphics[width = \textwidth]{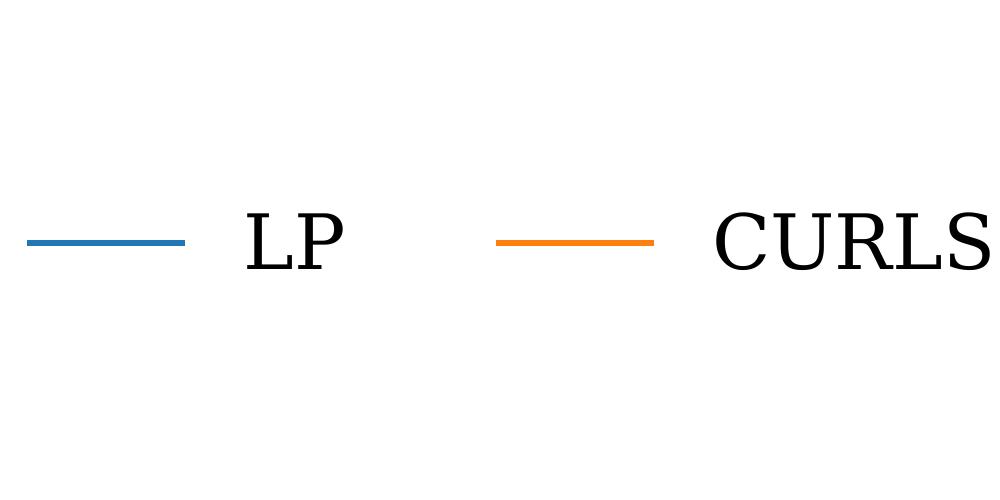} 
		\vspace{-10mm}
	\end{subfigure}
	\vspace{-4mm}
	\caption{Comparison of the \lcapm and LP solution to the \lcap with respect to normalized volume (left) and runtime (right).}
	\label{fig:relu-method-exp}
\end{wrapfigure}
\paragraph{Linear Constraint Aggregation}\label{sec:results-relu-method}
To evaluate \lcapm on the Linear Constraint Aggregation problem (\lcap), we compare it to an LP-based approach based on \cref{eq:LCAP} and implemented using a commercial LP solver (GUROBI \citep{gurobi}). However, considering all soundness constraints associated with the $2^d$ corner points is intractable. Therefore, we use an iterative sampling strategy (see \cref{sec:dpcap-baselines} for more details).

To compare the obtained relational constraints, we consider the volumes induced by the two methods and report mean normalized abstraction volumes $\vol^\text{LP}/ \vol^\text{\lcapm}$ in \cref{fig:relu-method-exp} for sets of $m=4$ randomly generated constraints in $d=[5,100]$ dimensions (see \cref{sec:DP-dataset} for more details). We observe that while the LP based solutions are more precise for up to 75 dimensional problems, they take around 5 orders of magnitude longer to compute. For higher dimensional problems, \lcapm is both faster and more precise. 
During the certification of a single input, we consider multiple hundred up to $512$ dimensional \lcap problems, making even the sampling based LP solution infeasible in practice and highlighting the importance of the efficient constraint aggregation via \lcapm for the \tool framework.
 
\section{Related Work}\label{sec:related-work}

\paragraph{Empirical Robustness of \nodes}
\citet{yan2019robustness} introduce TisODEs, by adding a regularization term to the loss which penalizes differences between neighboring trajectories to improve empirical robustness.
A range of work \citep{kang2021stable,rodriguez2022lyanet,huang2020adversarial,zakwan2022robust} trains NODEs which satisfy different forms of Lyapunov stability \citep{justus2008ecological}, yielding increased empirical robustness. However, \citet{huang2020adversarial} have shown that these empirical robustness improvements might be due to gradient obfuscation \citep{athalye2018obfuscated} caused by the use of adaptive step-size solvers.
Furthermore, \citet{carrara2022improving} have shown that varying the solver tolerance during inference can increase empirical robustness.

\paragraph{Verification and Reachability Analysis of \nodes}
\citet{lopez2022reachability} analyze the dynamics of very low dimensional ($d<10$) \nodes using CORA \citep{Althoff13} and the (polynomial) Zonotope domain, and those of higher dimensional linear \nodes using the star set domain. In contrast to our work, they analyze only the learned dynamics, excluding the solver behavior, which has a significant effect on practical robustness \citep{huang2020adversarial}.
\citet{grunbacher2021verification} introduce stochastic Lagrangian reachability to approximate the reachable sets of \nodes with high confidence by propagating concrete points sampled from the boundary of the input region.
However, the number of required samples depends exponentially on the dimension of the problem, making it intractable for the high-dimensional setting we consider.
\citet{huang2022fi} propose forward invariance ODE, a sampling-based verification approach leveraging Lyapunov functions.
\ccc{Moreover, when using fixed step size ODE solvers the verification of NODEs can be seen as verifying neural network dynamic models \citep{adams2022formal,wei2022safe} or by unrolling them even conventional feed-forward neural networks.}

\paragraph{Neural Network Verification} 
Deterministic neural network verification methods, typically either translate the verification problem into a linear \citep{PalmaBBTK21,MullerMSPV22,WangZXLJHK21,XuZ0WJLH21}, mixed integer \citep{TjengXT19,SinghGPV19}, or semidefinite \citep{RaghunathanSL18,DathathriDKRUBS20} optimization problem, or propagate abstract elements through the network \citep{singh2019abstract,gowal2019scalable,SinghGMPV18}
To obtain models amenable to certification, certified training \citep{MirmanGV18,GowalDSBQUAMK18,zhang2019towards} methods use the latter class of approaches to compute and optimize a worst-case over-approximation of the training loss. 
However, none of these methods support the analysis of NODEs without substantial extensions.
\section{Conclusion}
In this work, we propose the analysis framework \tool, \tooll, which, for the first time, allows the verification and certified training of high dimensional \nodes based on the following key ideas:
i) We introduce \solver solvers which retain the efficiency of adaptive solvers but are restricted to discrete instead of continuous step-sizes. 
ii) We leverage \solver solvers to construct efficient graph representations of all possible solver trajectories given an input region. 
iii) We build on linear bound propagation based neural network analysis and propose new algorithms to efficiently operate on these graph representations.
Combined, these advances enable \tool to analyze \nodes under consideration of solver effects in polynomial time.
\section{Ethics Statement}
As \tool, for the first time, enables the certified training and verification of \nodes, it could help make real-world AI systems more robust to both malicious and random interference. Thus any positive and negative societal effects these systems have already could be amplified. Further, while we obtain formal robustness guarantees for $\ell_\infty$-norm bounded perturbations, this does not (necessarily) indicate sufficient robustness for safety-critical real-world applications, but could give practitioners a false sense of security.

\section{Reproducibility Statement}
We publish our code, all trained models, and detailed instructions on how to reproduce our results at \url{https://github.com/eth-sri/GAINS} and provide an anonymized version to the reviewers. 
Further algorithmic details can be found in \cref{sec:latent_ode,app:node_trafo_details}.
Additionally, in \cref{app:details-experiments,app:classification-details,app:time-series-forecasting-details}
we provide implementation details and further discussions for our general method, classification tasks, and time-series forecasting tasks resistively.
Lastly, details on the adversarial attacks and \lcap dataset used in \cref{sec:trajectory-attacks} can be found \cref{app:trajectory-attack,app:dpcap_details} respectively.

\section*{Acknowledgements}
This work is supported in part by ELSA --- European Lighthouse on Secure and Safe AI funded by the European Union under grant agreement No. 101070617. Views and opinions expressed are however those of the authors only and do not necessarily reflect those of the European Union or European Commission. Neither the European Union nor the European Commission can be held responsible for them.

\message{^^JLASTBODYPAGE \thepage^^J}

\clearpage
\bibliography{references}
\bibliographystyle{iclr2023_conference}

\message{^^JLASTREFERENCESPAGE \thepage^^J}

\ifincludeappendixx
	\clearpage
	\appendix
	\section{Latent \odes for Time-Series Forcasting} \label{sec:latent_ode}

\begin{figure*}
	\centering
	\begin{tikzpicture} [ scale = 0.90,
    font= \scriptsize,
    >=LaTeX,
    transform shape,
    font = \normalsize,
    hidden-enc/.style={%
        rectangle, 
        draw,
        minimum width=4mm,
        minimum height=4mm,
        fill=green,
        rotate = 45,
        },
    decoder-init/.style={%
        circle, 
        draw,
        radius = 2.3mm,
        fill=red,
        },
    hidden-dec/.style={%
        rectangle, 
        draw,
        minimum width=4mm,
        minimum height=4mm,
        fill=red,
        rotate = 45,
        },
    point/.style={%
        circle, 
        draw,
        radius = 2pt,
        fill=blue,
        inner sep = -2.,
        },
    point2/.style={%
        circle, 
        draw,
        radius = 2pt,
        fill=cyan,
        inner sep = -2.,
        },
    box/.style={%
        rectangle, 
        draw,
        minimum width=4mm,
        minimum height=8mm,
        fill=orange,
        },
    ArrEnc/.style={%
        rounded corners=.2cm,
        line width = 0.5mm,
        green,
        },
    ArrDec/.style={%
        rounded corners=.2cm,
        line width = 0.5mm,
        red,
        },
    ArrE1/.style={%
        line width = 0.25mm,
        },
    Dashed/.style={%
        dashed,
        line width = 0.25mm,
        },
    ArrowC2/.style={%
        rounded corners=.5cm,
        thick,
        },
    ]
    
    \pgfmathsetmacro{\shift}{0}
    \node[hidden-enc](h1) at (0.4 +\shift,0) {}; 
    \node[hidden-enc](h2) at (1.5 +\shift,-1) {}; 
    \node[hidden-enc](h3) at (3 + \shift,-0.25) {}; 
    \node[hidden-enc](h4) at (4.3+ \shift,-0.9) {}; 
    \draw[-,ArrEnc] (h1.south east) --(0.75,-0.2)-- (1.25,0.05)-- (1.5+ \shift,-0.3);
    \draw[->,ArrE1] (1.5+ \shift ,-0.35) -- (h2.north east);
    
    \draw[-,ArrEnc] (h2.south east) --(1.95,-0.6)-- (2.5,-1.2)-- (3+ \shift,-0.9);
    \draw[->,ArrE1] (3+ \shift,-0.85) -- (h3.south west);
    
    \draw[-,ArrEnc] (h3.south east) --(3.45+ \shift,-0.1)-- (3.65+ \shift,0.1) -- (3.9+ \shift,0.2)-- (4.3+ \shift,-0.3);
    \draw[->,ArrE1] (4.3+ \shift,-0.35) -- (h4.north east);

    \node[point] (p1) at (4.3 + \shift,{0.5*sin(0.4 r )-2.25}) {};
    \node[] (l1) at (4.1 + \shift,{0.5*sin(0.4 r )-2.55}) {$\vx_{1}$};
    \draw[->,Dashed] (p1) -- (4.3 + \shift, -1.3);
    
    \node[point] (p2) at (3 + \shift,{0.5*sin(1.5 r )-2.3}) {};
    \node[] (l2) at (2.8 + \shift,{0.5*sin(1.5 r )-2.55}) {$\vx_{2}$};
    \draw[->,Dashed] (p2) -- (3.0 + \shift, -1.1);
    
    \node[point] (p3) at (1.5 + \shift,{0.5*sin(3.0 r )-2.25}) {};
    \draw[->,Dashed] (p3) -- (1.5 + \shift, -1.35);
    
    \node[point] (p5) at (0.4 + \shift,{0.5*sin(4.3 r )-2.2}) {};
    \draw[->,Dashed] (p5) -- (0.4 + \shift, -0.5);
    \node[] (l5) at (0.2 + \shift,{0.5*sin(4.3 r )-2.5}) {$\vx_{L'}$};

    \draw[->, line width=0.5mm] (4.7+ \shift,-3.3) -- (-0.5+\shift,-3.3);
    \draw[-,line width=0.2mm] (0.4+\shift,-3.45) -- (0.4+\shift,-3.15);
    \node[] (t1) at (0.4 + \shift,-3.65) {$t_{L'}$};
    \draw[-,line width=0.2mm] (1.5+\shift,-3.45) -- (1.5+\shift,-3.15);
    \node[] (t2) at (3. + \shift,-3.65) {$t_{2}$};
    \draw[-,line width=0.2mm] (3+\shift,-3.45) -- (3+\shift,-3.15);

    \draw[-,line width=0.2mm] (4.3+\shift,-3.45) -- (4.3+\shift,-3.15);
    \node[] (t5) at (4.3 + \shift,-3.65) {$t_{1}$};
    
    \node[](g1) at (2.0 +\shift,-0.25) {\footnotesize GRU};
    \node[](n1) at (2.25 +\shift,-1.65) {\footnotesize NODE};
    
    \node[box,label={$p(\vz|\vx_{ts}^{L'})$},align=center] (b1) at (5.5,-0.9) {$\bs{\mu}$ \\ $\bs{\sigma}$};
    \draw[->,thick,orange] (h4) -- (b1); 
    
    \node[] (sim) at (6.2,-0.9) {\large $\sim$};

    \pgfmathsetmacro{\shift}{7.3}
    \node[decoder-init,label = {$\vz(0)$}](y0) at (-0.5 +\shift,-0.9) {};
    \node[hidden-dec](y1) at (0.4 +\shift,-0.4) {}; 
    \node[hidden-dec](y2) at (1.5 +\shift,-0.0) {}; 
    \node[hidden-dec](y3) at (3 + \shift,-0.7) {}; 
    \node[hidden-dec](y4) at (4.3+ \shift,-1.2) {}; 
    \node[hidden-dec] (y5) at (6.3 +\shift,-0.6) {}; 
    \node[](yy) at (0.37+ \shift, 0.25) {$\vz(t_{1})$};
    \node[](yy) at (6.35+ \shift, -0.05) {$\vz(t_{L})$};
    \draw[-,ArrDec] (y0.east) --(0 + \shift, -0.8)-- (y1.west);
    \draw[-,ArrDec] (y1.east) --(1. + \shift, 0.3)-- (y2.north);
    \draw[-,ArrDec] (y2.south) -- (1.8 + \shift, -0.6)--(2.5 + \shift, -0.2)-- (y3.north);
    \draw[-,ArrDec] (y3.south) -- (3.6 + \shift, -1.55) -- (y4.west);
    \draw[-,ArrDec] (y4.east) --(4.9 + \shift, -0.2)--(5.2 + \shift, 0.)-- (5.7 + \shift, -0.25) -- (y5.north);

    \node[point2] (pp1) at (0.4 + \shift,{0.5*sin(0.4 r )-2.25}) {};
    \node[] (l1) at (0.2 + \shift,{0.5*sin(0.4 r )-2.55}) {$\hat{\vx}_{1}$};
    \draw[->,Dashed] (y1.south west) -- (0.4 + \shift, -1.8) ;
    \draw[->,Dashed] (y2.south west) -- (1.5 + \shift, -1.6) ;
    \draw[->,Dashed] (y3.south west) -- (3.0 + \shift, -1.9) ;
    \draw[->,Dashed] (y4.south west) -- (4.3 + \shift, -2.4) ;
    \draw[->,Dashed] (y5.south west) -- (6.3 + \shift, -1.95) ;
    
    \node[point2] (pp2) at (1.5 + \shift,{0.5*sin(1.5 r )-2.3}) {};
    \node[] (l2) at (1.2 + \shift,{0.5*sin(1.5 r )-2.55}) {$\hat{\vx}_{2}$};
    \node[point2] (pp3) at (3.0 + \shift,{0.5*sin(3.0 r )-2.25}) {};
    \node[point2] (pp5) at (4.3 + \shift,{0.5*sin(4.3 r )-2.2}) {};
    \node[] (l5) at (4.1 + \shift,{0.5*sin(4.3 r )-2.5}) {$\hat{\vx}_{L'}$};
    \node[point2] (pp6) at (6.3 + \shift,{0.5*sin(6.3 r )-2.25}) {};
    \node[] (l5) at (5.8 + \shift,{0.5*sin(6.3 r )-2.55}) {$\hat{\vx}_{L}$};
    
    \draw[->, line width=0.5mm] (-0.9+\shift,-3.3) -- (7.25+ \shift,-3.3);
    \draw[-,line width=0.2mm] (-0.5+\shift,-3.45) -- (-0.5+\shift,-3.15);
    \node[] (t0) at (-0.5 + \shift,-3.65) {$0$};
    \draw[-,line width=0.2mm] (0.4+\shift,-3.45) -- (0.4+\shift,-3.15);
    \node[] (t1) at (0.4 + \shift,-3.65) {$t_{1}$};
    \draw[-,line width=0.2mm] (1.5+\shift,-3.45) -- (1.5+\shift,-3.15);
    \node[] (t2) at (1.5 + \shift,-3.65) {$t_{2}$};
    \draw[-,line width=0.2mm] (3+\shift,-3.45) -- (3+\shift,-3.15);
    
    \draw[-,line width=0.2mm] (4.3+\shift,-3.45) -- (4.3+\shift,-3.15);
    \node[] (t5) at (4.3 + \shift,-3.65) {$t_{L'}$};
    \draw[-,line width=0.2mm] (6.3+\shift,-3.45) -- (6.3+\shift,-3.15);
    \node[] (t6) at (6.3 + \shift,-3.65) {$t_{L}$};
    
    \node[](n2) at (2.15 +\shift,-0.8) {\footnotesize NODE};
    
    \node[](e) at (2.5,.8) {\large Encoder $\ve_{\bs{\theta}}$};
    \node[](e) at (2.5 +\shift,.8) {\large Decoder $\vd_{\bs{\theta}}$};

\end{tikzpicture} %
	\caption{Visualization of the latent ODE with ODE-RNN encoder. Due to the NODE layer in the decoder the model is able to estimate the data point of the time-series at any desired time. Figure inspired by \citep{chen2018neural,rubanova2019latent}.}
	 \label{fig:latent-ode}
\end{figure*}
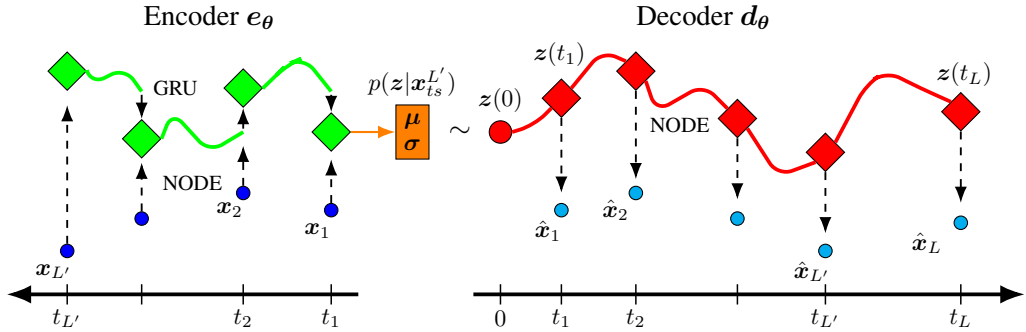

For time-series forecasting, we use an encoder-decoder architecture called latent ODE \citep{rubanova2019latent} and illustrated in \cref{fig:latent-ode}. 
The encoder $\ve_{\bs{\theta}}$ is an ODE-RNN, yielding an embedding $\bm{s}_{L'}$ of the data points observed until $t_{L'}$, where the series is processed in reversed time order. The core idea is to describe the evolution of a hidden state with a NODE and update it using a GRU unit \citep{cho2014properties} (described in \cref{app:gru-update}) to account for new observations. This embedding is then passed through a one layer MLP to yield the posterior distribution $p(\vz|\vx_{ts}^{L'}) = \bc{N}(\bm{\mu}, \bm{\sigma})$ over the initial state of the decoder $\vz(0)$.
The decoder $\vd_{\bs{\theta}}$ then estimates $\hat{\vx}_{L}$ as a linear transform of the solution $\vz(t_{L})$ of the IVP with initial state $\vz(0)$ at time $t_L$.
Note that in testing we use $\vz(0) = \bm{\mu}$ and omit the sampling.

The latent ODE is trained to maximize the evidence lower bound (ELBO) \citep{kingma2013auto} an minimize the absolute error of the final predictions weighted with $\gamma$:
\begin{align} 
\label{eq:forecasting-loss-background}
\mathcal{L}_f(\vx^L_{ts},L') &= \gamma \cdot \lVert \hat{\vx}_{L} - \vx_{L}\rVert_1 -  \ELBO(\vx^L_{ts},L') \\
\ELBO(\vx^L_{ts},L') &= \mathbb{E}_{\vz' \sim p_{\mathcal{N}}}\left[\log\left(\vd_{\bs{\theta}}\left(\vz', t_{L}\right)\right)\right] - \KL\left[p_{\mathcal{N}}|| p\right].
\end{align}

\subsection{GRU update} \label{app:gru-update}
In \cref{fig:gru-cell} we show the update of the hidden state $\bm{s}_{i-1}$ of the \ode-\RNN~\citep{rubanova2019latent} architecture after feeding the $i$-th entry $(\vx_{i},t_{i})$ as input.
The update uses a \node layer to represent $f_{z}$, where the integration domain of the \node layer is $[t_{i-1}, t_{i}]$.
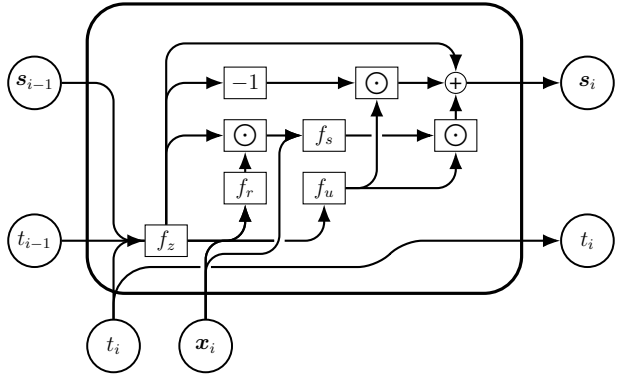
\begin{figure*}
    \centering
    \begin{tikzpicture}[scale=0.7,transform shape,
        font= \large,
        >=LaTeX,
        cell/.style={%
            rectangle, 
            rounded corners=5mm, 
            draw,
            very thick,
            },
        operator/.style={%
            circle,
            draw,
            inner sep=-0.5pt,
            minimum height =.4cm,
            },
        function/.style={%
            ellipse,
            draw,
            inner sep=1pt
            },
        ct/.style={%
            circle,
            draw,
            line width = .75pt,
            minimum width=1cm,
            inner sep=1pt,
            },
        gt/.style={%
            rectangle,
            draw,
            minimum width=8mm,
            minimum height=6mm,
            inner sep=1pt
            },
        ArrowC1/.style={%
            rounded corners=.25cm,
            thick,
            },
        ArrowC2/.style={%
            rounded corners=.5cm,
            thick,
            },
        ]
    
        \node [cell, minimum height =5.5cm, minimum width=8.25cm] at (1.125,0.25){} ;
        \node [gt] (fz) at (-1.5,-1.5) {$f_z$};
        \node [gt] (fr) at (0,-0.5) {$f_r$};
        \node [gt] (fu) at (1.5,-0.5) {$f_u$};
        \node [gt] (fs) at (1.5, 0.5) {$f_s$};
    
        \node [gt] (mux1) at (-0,0.5) {$\bigodot$};
        \node [gt] (mux2) at (4,0.5) {$\bigodot$};
        \node [gt] (mux3) at (2.5,1.5) {$\bigodot$};
        \node [gt] (neg) at (0,1.5) {$-1$};
        \node [operator] (add1) at (4,1.5) {+};
    
        \node[ct] (t) at (-4,-1.5) {$t_{i-1}$};
        \node[ct] (t2) at (-2.5,-3.5) {$t_{i}$};
        \node[ct] (z) at (-4,1.5) {$\vs_{i-1}$};
        \node[ct] (x) at (-0.75,-3.5) {$\vx_{i}$};
    
        \node[ct] (t3) at (6.5,-1.5) {$t_{i}$};
        \node[ct] (z2) at (6.5,1.5) {$\vs_{i}$};

        \draw [ArrowC1] (t2) -- (t2 |- t)-- (fz);
        \draw [ArrowC1] (z) -- (-2.5,1.5) -- (-2.5,1.5 |- t)-- (fz);
        \draw [->,ArrowC1] (t) -- (fz);
        \draw [->,ArrowC1] (x) -- (-0.75,-1.5) -- (0,-1.5)-- (fr);
        \draw [->,ArrowC1] (fz) -- (0,-1.5)-- (fr);
        \draw [-,ArrowC1] (x) -- (-0.75,-1.5) -- (0.55,-1.5);
        \draw [->,ArrowC1] (0.75,-1.5) -- (1.5,-1.5)-- (fu);
        \draw [-,ArrowC1] (fz) -- (0.55,-1.5);
        
        \draw[->,ArrowC1] (fz) -- (-1.5,0.5) -- (mux1);
        \draw[->,ArrowC1] (fr) -- (mux1);
         
         \draw[->,ArrowC1] (mux1.east) --  (fs.west);
         \draw [->,ArrowC1] (x) -- (-0.75,-1.75)-- (0.65,-1.75) -- (0.65,0.5) -- (fs.west);
         
         \draw [->,ArrowC1] (fu) -- (2.5,-0.5) --(mux3);
         \draw[->,ArrowC1] (fz) -- (-1.5,1.5) -- (neg);
         \draw[->,ArrowC1] (neg)--(mux3);
         \draw[->,ArrowC1] (mux3) -- (add1);
         
         \draw[->,ArrowC1] (mux2) -- (add1);
         \draw[->,ArrowC1] (fu) -- ( 4,-0.5)--(mux2);
         \draw[-,ArrowC1] (fs) -- (2.4,0.5);
         \draw[->,ArrowC1] (2.6,0.5) -- (mux2);
         \draw[->,ArrowC1] (fz) -- (-1.5,2.25) -- (4,2.25) --(add1);
         
        \draw [-,ArrowC2] (t2) |- (-0.85,-2);
        \draw [->,ArrowC1]  (-0.65,-2) -- (2.5,-2)-- (3,-1.5) -- (t3);
        \draw[->,ArrowC1] (add1) -- (z2) ;
    
    \end{tikzpicture}
    \caption{GRU-update for the \ode-\RNN architecture, where $\odot$ denotes the hadamard product (componentwise multiplication) of two vectors and $f_{z}, \ f_{u},\ f_{r},\ f_{s}$ are auxiliary NNs. } \label{fig:gru-cell}
\end{figure*}

\section{Provable \node Training} \label{app:node_trafo_details}
In this section, we describe our \tool{}-based training procedure.
We consider the setting with data distribution $(\vx, y) \sim \mathcal{D}$ and we compute the \node input $\vz_0$ (either $\vz_0 := \vx$ or via some encoder) with the corresponding bounds $\bc{Z}$.
Standard provable training aims to optimize a loss based on the over-approximation (\cref{eq:rob_opt}). However, in the case of \node it is intractable to compute the full over-approximation of the trajectory graph (discussed in \cref{sec:verification}) for each sample in training. Thus,  we only sample up to $\kappa$ selected trajectories from $\bc{G}(\bc{Z})$.

\paragraph{Trajectory Exploration}
During the sampling we balance exploration of the full trajectory graph and staying close to the reference trajectory, the trajectory $\Gamma(\vz_0)$ of the solver with unperturbed input $\vz_0$.
A visualization of the selection process is depicted in \cref{fig:flow-chart}.

We select trajectories as follows:
We start the propagation of $\bc{Z}$ through the \node layer.
Recall that, for a concrete input at each step the \solver solver will either \emph{(i) increase}, \emph{(d) decrease} or \emph{(a) accept}, i.e., keep, the current step size $h$. For an abstract solver step we may need to keep track of multiple decisions (trajectory splitting). 
Thus, for each abstract solver step we check whether or not trajectory splitting occurs and as long as no trajectory split occurs, we are following the reference trajectory.
If, however, multiple updates are possible, i.e., we encounter trajectory splitting, we choose a single path $u$ via random sampling (details below), and add the corresponding state to the branching point set $\mathcal{C}$.
Afterward, we check whether or not we have reached $T_{end}$, where if $T_{end}$ is reached, we save the resulting trajectory to a set $\mathcal{S}$.
Moreover, we repeat the process with a checkpoint $C \in \mathcal{C}$, as long as there is still a checkpoint in $\mathcal{C}$, i.e. $|\mathcal{C}| > 0$, and we have not already collected $\kappa$ trajectories, i.e. $|\mathcal{S}| < \kappa$.

\paragraph{Sampling Updates}
For a state $(t, h$) we let $V_{(t,h)}$ denote the set of vertices which where traversed from initial vertex $(0,\ h_0)$ to $(t, \ h)$.
Moreover, for any vertex $v = (\tilde{t},\ \tilde{h})$ we define its reference vertex $v' = (\tilde{t}', \ \tilde{h}')$ as the vertex with the smallest $\ell_1$-distance to the vertex $v$ among the vertices in the reference trajectory $\Gamma(\vz_0)$, i.e.
\begin{align}
    v' = (\tilde{t}', \ \tilde{h}') = \argmin_{(\hat{t},\hat{h}) \in \Gamma(\vz_0)} |\tilde{t} -\hat{t}| + |\tilde{h} -\hat{h}|.
\end{align}
Furthermore, for any vertex $v \in V_{(t,h)}$ we let $u(v)$ denote the update (\emph{(i) increase}, \emph{(d) decrease} or \emph{(a) accept}) taken to leave state $v$ in the given trajectory.
Analogously, we define for any $v' \in \Gamma(\vz_0)$ $u'(v')$ as the performed update in $\Gamma(\vz_0)$ after vertex $v'$.

Additionally, we define the auxiliary mapping $g_{n}:\{\text{\emph{d}, \emph{a}, \emph{i}}\} \to \{0,\ 1,\ 2\}$, where $g_{n}(d) = 0, \ g_{n}(a) = 1$ and $g_{n}(i) = 2$. 
Using the previous definitions we define the location index of $V_{(t,h)}$ as $n(V_{(t,h)}) = \sum_{v \in V_{(t,h)}} g_{n}(u(v)) - g_{n}(u'(v'))$.
If the location index is bigger than zero, we assume to be traversing a trajectory that has performed steps with bigger step sizes than the reference trajectory $\Gamma(\vz_0)$.
On the other hand, for a location index smaller than zero the opposite is true, whereas if the location index is zero we are close to the reference trajectory $\Gamma(\vz_0)$.

Finally, when sampling an update $u$ we choose from the categorical distribution $P_u(p_d,p_a,p_i)$ depending on  $n(V_{(t,h)}),u'(v')$ for the current state $(t,h)$ and hyperparameters $q_1$ and $q_2$.
The definition of the probabilities $p_d, \ p_a$ and $p_i$ can be seen in \cref{tab:sampling}.

In the definition of the sample probabilities the update that pushes the location index the most towards zero occurs always with probability $1-q_1 -q_2$, whereas the event occurring with probability $q_1$ pushes the location index away from zero.
Hence, depending on which probability is higher, we either prefer to select trajectories close to the reference trajectory or trajectories that are distributed over the entire trajectory graph.
In order to have a combination of both, we use an annealing process for the hyperparameters $q_1$ and $q_2$.
In the early stages of training, we choose selection hyperparameters such that $1-q_1 -q_2 \geq q_2 \geq q_1$, i.e. stay close to the reference trajectory, and towards the end of the training the chain of inequalities should be reversed, i.e. cover the entire trajectory graph and not just a region.

\paragraph{Checkpoint Selection Criterion}
We use the following decision criterion to select $C^*$ from $\mathcal{C}$
\begin{equation}\label{eq:decision-criteria}
    C^* = \argmax_{C = \{V_C\} \in \mathcal{C}} \frac{|n(V_C) - n_\mathcal{S}|}{2} - |V_C| - \sigma_{out}[V_C],
\end{equation}
where the vertex set $V_C$ contains all traversed vertices until the creation of the checkpoint $C$ and we denote by $n_{\mathcal{S}}$ the average location index of the already stored trajectories in $\mathcal{S}$.
Observe, that the decision criterion is designed such that checkpoints in under-explored regions of the trajectory graph and checkpoints arising early in the trajectory graph are favored, where the former statement is captured by the first term in \cref{eq:decision-criteria}, whereas the remaining two terms capture the latter statement.

\begin{table}[ht]
\caption{The definition of the probabilities $p_d, \ p_a$ and $p_i$ depending on the location index $n(V)$, reference update ${u'}$ and hyperparameters $q_1, \ q_2$.}
  \label{tab:sampling}
\centering
\begin{tabular}{ccccc}
\toprule
	$n(V)$       & $u'$                                 & $p_d$ & $p_a$ & $p_i$                                           \\
\midrule
	$n = 0$ & \emph{a}                                  & $\frac{q_1 + q_2}{2}$ & $1-q_1 - q_2$ & $\frac{q_1 + q_2}{2}$ \\
\cmidrule(l){1-5}
	$n = 0$ & \emph{d}                                  & \multirow{2}{*}{ $1-q_1 - q_2$} & \multirow{2}{*}{$q_2$} & \multirow{2}{*}{$q_1$}\\
	$n > 0$ & $\{\text{\emph{d}, \emph{a}, \emph{i}}\}$ &  & & \\
\cmidrule(l){1-5}
	$n = 0$ & \emph{i} & \multirow{2}{*}{$q_1$} & \multirow{2}{*}{$q_2$} & \multirow{2}{*}{$1 - q_1 - q_2$}\\
	$n < 0$ & $\{\text{\emph{d}, \emph{a}, \emph{i}}\}$ &   & &  \\
\bottomrule
\end{tabular}
\end{table}

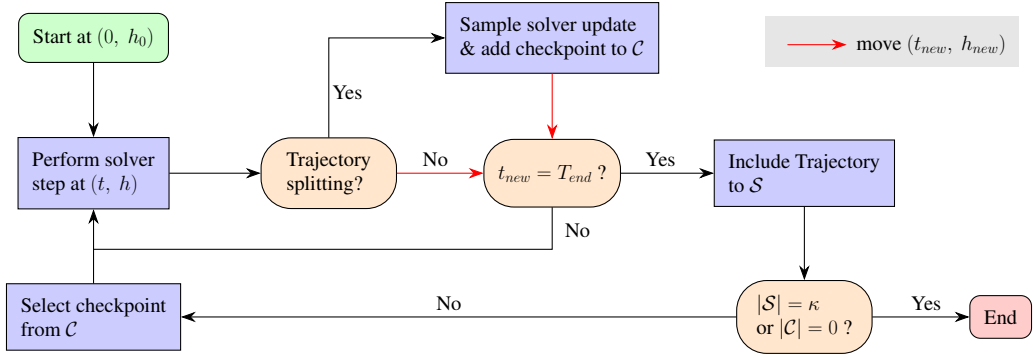
\begin{figure}
	\centering
	\begin{tikzpicture}[node distance = 2cm,scale=0.45, transform shape,inner sep = 12., font=\LARGE, decision2/.style={%
		rectangle, 
		draw,
		minimum width=40mm,
		minimum height= 20mm,
		fill=orange!20,
		align = left,
		rounded corners = 0.4cm,
		}]
		
	\node [terminator, fill=green!20] at (0,2) (start) {Start at $(0,\ h_0)$};
	\node [process, fill=blue!20] at (0,-2) (p1) {Perform solver \\ step  at $(t,\ h)$};
	
	\node [decision2] at (6.95,-2.0) (d1) { Trajectory \\ splitting?};
	\node [decision2] at (13.55,-2) (d2) { $t_{new} = T_{end}$ ?};
	\node [process, fill=blue!20] at (13.55,2) (p2) {Sample solver update \\ \& add checkpoint to $\mathcal{C}$};
	\node [process, fill=blue!20] at (21,-2) (p3) {Include Trajectory \\ to $\mathcal{S}$};
	\node [decision2] at (21,-6.25) (d3) { $|\mathcal{S}| = \kappa$ \\or $|\mathcal{C}| = 0$ ?};
	\node [process, fill=blue!20] at (0,-6.25) (p4) {Select checkpoint \\ from $\mathcal{C}$};
	\node [terminator, fill=red!20] at (26.8,-6.25) (end) {End};
	
	\draw[-{Stealth[length=2mm, width=1.4mm]}] (p1.east) -- (d1.west);
	
	\draw[-{Stealth[length=2mm, width=1.4mm]},red] (d1.east) -- (d2.west);
	\draw[-{Stealth[length=2mm, width=1.4mm]}] (d1.north) -- (6.95,2)--(p2.west);
	\draw[-{Stealth[length=2mm, width=1.4mm]},red] (p2.south) --(d2.north);
	
	\draw[-{Stealth[length=2mm, width=1.4mm]}] (d2.east) -- (p3.west);
	\draw[-] (d2.south) -- (13.55,-4.25)-- (0,-4.25);
	
	\draw[-{Stealth[length=2mm, width=1.4mm]}] (p3.south) -- (d3.north);
	\draw[-{Stealth[length=2mm, width=1.4mm]}] (d3.west) -- (p4.east);

	\draw[-{Stealth[length=2mm, width=1.4mm]}] (p4.north) -- (p1.south);
	\draw[-{Stealth[length=2mm, width=1.4mm]}] (start.south) -- (p1.north);
	\draw[-{Stealth[length=2mm, width=1.4mm]}] (d3.east) -- (end.west);
	
	\node[rectangle, fill = gray!20, minimum width = 75mm, minimum height = 15mm] at (23.6,1.9) (legend) {};
	
	\draw[-{Stealth[length=2mm, width=1.4mm]},red] (20.5,1.8) -- (22.25,1.8);
	\node[align = left] at (24.75,1.8) (l1) {move $(t_{new},\ h_{new})$};
	
	\node[draw=none] at (10.1, -1.6) (no1) {No};
	\node[draw=none] at (10.5,-5.85) (no2) {No};
	\node[draw=none] at (14.3,-3.7) (no3) {No};
	\node[draw=none] at (16.75, -1.6) (yes) {Yes};
	\node[draw=none] at (24.6, -5.85) (yes1) {Yes};
	\node[draw=none] at (7.5, 0.4) (yes1) {Yes};

	\end{tikzpicture} 
	\caption{Selection process of $\mathcal{S}$, which contains at most $\kappa$ trajectories starting with initial step size $h_0$ and final integration time $T_{end}$ and the branching point set $\mathcal{C}$.}
	\label{fig:flow-chart}
\end{figure}

\paragraph{Loss Computation}
Finally, we compute the \boxd output of the \node layer as the over-approximation of the final states form all saved trajectories $\mathcal{S}$.
Then, for provable training we use a loss term of the following form:
\begin{align} \label{eq:cert-loss}
\bc{L}(\vz_0, \bc{Z}, y) &= (1 - \sfrac{\omega_1\epsilon'}{\epsilon_t}) \bc{L}_\text{std}(\vz_0,y) +  \sfrac{\omega_1\epsilon'}{\epsilon_t} \bc{L}_\text{rob}(\bc{Z},y) + \omega_2 \|\vu_\text{out}-\vl_\text{out}\|_1,
\end{align}
where $\bc{L}_\text{std}$ is the standard loss (depending on the task) evaluated on the unperturbed sample, and $\bc{L}_\text{rob}$ is an over-approximation of $\bc{L}_\text{std}$ based on the abstraction obtained from $\mathcal{S}$.
The term $\vu_\text{out}-\vl_\text{out}$ regularizes the bound width of the corresponding output region.
During training, we anneal $\epsilon$, gradually increasing $\epsilon'$ from $0$ to $\epsilon$, thereby shifting focus from the standard to the robust loss term. 
In the classification setting, we use the cross entropy loss and in time series forecasting we use a latent ODE specific loss, combining a MAE error and ELBO term, defined in \cref{eq:forecasting-loss-background}.
\newpage
\begin{wrapfigure}[12]{r}{0.48\textwidth}
		\vspace{-2mm}
		\begin{subfigure}[b]{0.9\linewidth}
			\centering
			\begin{tikzpicture}[transform shape,scale = 0.30,node distance = 66mm, xshift=-0.5]
				\node[] (sol) at (0,0) {
					\includegraphics[]{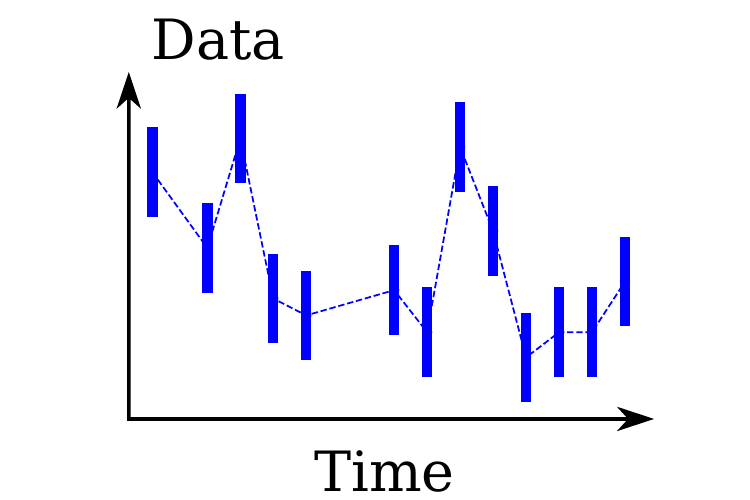} 
				};  
				\node[left of = sol ] (solver) {
						\includegraphics[]{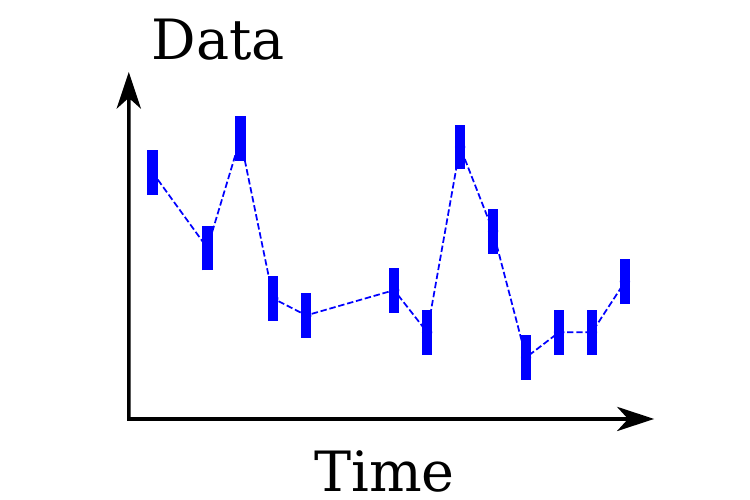} 
					};
				\node[left of = solver] {
						\includegraphics[]{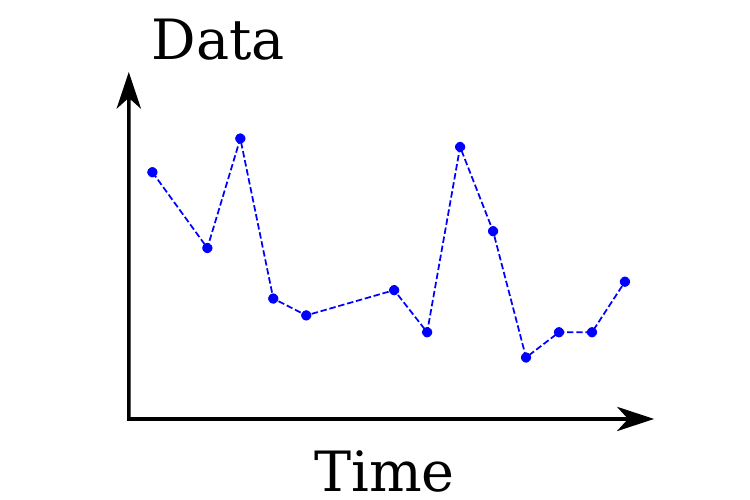} 
					};
			\end{tikzpicture}
			\vspace{-5mm}
			\caption{Annealing  perturbation $\epsilon$.}
			\label{fig:eps-annealing}
		\end{subfigure}
		\begin{subfigure}[b]{0.9\linewidth}
			\centering
			\begin{tikzpicture}[transform shape,scale = 0.30,node distance = 66mm]
				\node[] (sol2) at (0,0) {
					\includegraphics[]{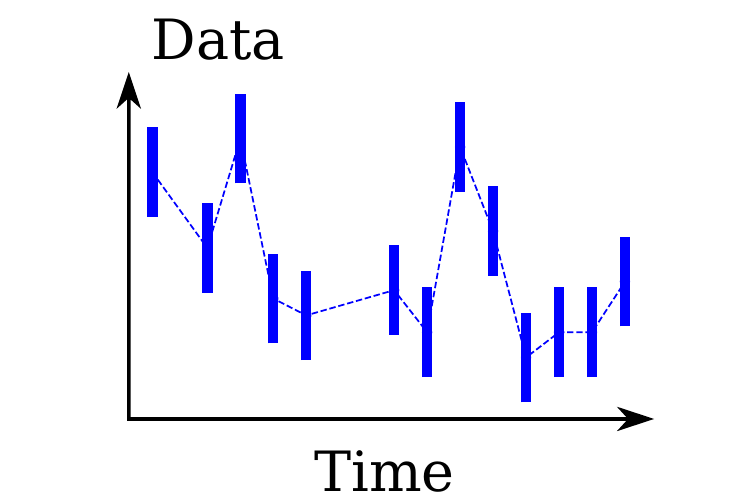} 
				};  
			\node[left of = sol2 ] (sol) {
					\includegraphics[]{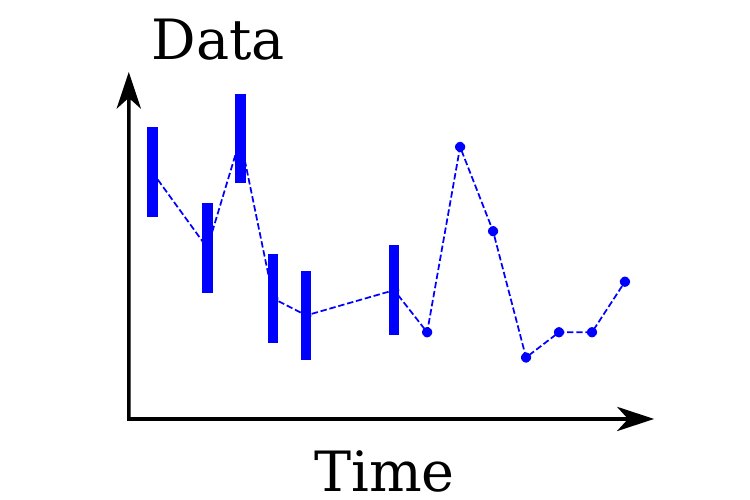} 
				};
			\node[left of = sol ] (solver) {
					\includegraphics[]{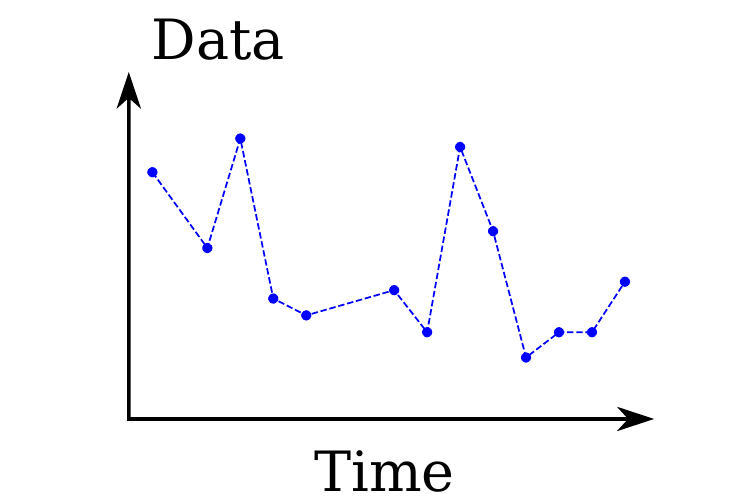} 
				};
			\end{tikzpicture}
			\vspace{-5mm}
			\caption{Annealing abstract ratio $\rho$.}
			\label{fig:rho-annealing}
		\end{subfigure}
		\vspace{-2mm}
		\caption{$\epsilon$-annealing (top) and $\rho$-annealing (bottom) for time-series input. Dots indicate concrete inputs and error bars abstract regions.}
		\label{fig:annealing}
		\end{wrapfigure}
\paragraph{Stabilizing Training}
In the time-series forecasting setting, the long integration times involving many solver calls lead to very large effective model depths. There, $\epsilon$-annealing alone is insufficient to stabilize the training in the face of an exponential accumulation of approximation errors.
To combat this, we additionally anneal the abstract ratio $\rho$ from $0$ to $1$ and only use non-zero perturbation magnitudes for the first $\rho L$ data points in every time series, i.e., for an input with time index $j$, we set $\epsilon' \gets \epsilon' \mathds{1}_{j \leq \rho L} $. We visualize this annealing process in \cref{fig:annealing} and highlight, that it is independent of $\epsilon$-annealing.
	
\ccc{\paragraph{Complexity Derivation}
The time complexity is derived via the maximum number of vertices in the trajectory graph $\bc{G}(\bc{Z})$.
Note that the graph is constructed using a CAS with update factor $\alpha$ that enforces a minimum step size $h_{min}$ (described in \cref{app:adaptive_solvers}).
The complexity does depend on $h_{min}$ and $\alpha$, but we consider both to be constant and have thus dropped the dependence.
We organize the graph into rows corresponding to the step sizes and observe that for integer $\alpha$ each step size contains at most $T_{end} / h_{min}$ vertices.
Further, the largest possible step size is $T_{end}$ and the smallest step size $h_{min}$.
Due to the exponentially spaced grid of possible step sizes with growth rate $\alpha$, it follows that the graph has at most $(\log(T_{end}) - log(h_{min}))/log(\alpha)$ different step sizes and hence rows.
Consequently there are at most $T_{end}/h_{min} (\log(T_{end}) - \log(h_{min}))/log(\alpha)$ or after dropping the constants $\mathcal{O}(T_{end} \log(T_{end}))$ vertices in $\bc{G}(\bc{Z})$.

For the final result, note that a simple graph with $v$ vertices has at most $v (v-1)/2$ edges.
Therefore, since all edges in the trajectory graph $\bc{G}(\bc{Z})$ represent a solver step, it follows that at most $\bc{O}(T_{end}^2 log^2(T_{end}))$ solver steps need to be considered by \tool.
}

\section{Experimental Details} \label{app:details-experiments}
We have used the \ode solvers from the torchdiffeq package\footnote{\hyperlink{https://github.com/rtqichen/torchdiffeq}{https://github.com/rtqichen/torchdiffeq}}~\citep{chen2018neural}, where we have extended the package to contain controlled adaptive \ode solvers. 
Moreover, we have used the PGD adversarial attack from the torchattacks package\footnote{\hyperlink{https://github.com/Harry24k/adversarial-attacks-pytorch}{https://github.com/Harry24k/adversarial-attacks-pytorch}}~\citep{kim2020torchattacks}.
The annealing processes of the perturbation $\epsilon$ use the implementation of the smooth scheduler from\footnote{\hyperlink{https://github.com/KaidiXu/auto_LiRPA/blob/master/auto_LiRPA/eps_scheduler.py}{https://github.com/KaidiXu/auto\_LiRPA/blob/master/auto\_LiRPA/eps\_scheduler.py}}~\citet{xu2020automatic}, which we denote as $\text{Smooth}(\epsilon_t,e_{start}, e_{end}, \text{mid})$. 
The first three arguments of the Smooth scheduler represent the target perturbation, the starting epoch of the scheduler, and the epoch in which the process reaches the target perturbation. 
The additional $\text{mid}$ parameter of the schedule is fixed to $\text{mid} = 0.6$ and anything else is used unaltered. 

Moreover, we use the annealing process $\text{Sin}(q_{start},q_{end},e_1, e_2)$,
for the hyperparameters $q_1, \ q_2$ occurring in the sampling process of the construction of the selection set $\mathcal{S}$ in \cref{app:node_trafo_details}.
The value $q$ of the annealing process $\text{Sin}(q_{start},q_{end},e_1, e_2)$ in epoch $e$ is given by

\begin{equation}
    q \gets\begin{cases}
q_{start},  &\text { if } e \leq e_{1}, \\
\sin\left(\pi\frac{e- e_{mid}}{e_2 - e_1}\right) \cdot \frac{q_{end} -q_{start}}{2} + \frac{q_{end} + q_{start}}{2},  \ &\text { else if }  e_1 < e \leq e_2, \\
q_{end}, &\text {otherwise,}
\end{cases}
\end{equation}
where we use $e_{mid} = \frac{e_2 + e_1}{2}$.

\subsection{\solver Details}  \label{app:adaptive_solvers}
When using a \solver, we have used in all experiments update factor $\alpha = 2$, momentum factor $\beta = 0.1$, absolute error tolerance $\atol = 0.005$ and the individual \ode solver steps where performed using the dopri5 \citep{dormand1980family} solver.
Additionally, we have introduced a minimal allowed step size constraint and a maximal number of allowed rejections after clipping for the \solver, where the minimum step size is fixed to $h_{min} = 0.02$ and the maximal number of allowed rejections after clipping is 2.
In our experiments on the \mnist, \fmnist, and \physio datasets the constraints only became active in early stages of training.
Note that only after rejecting a step with step size $h$ the aforementioned events can occur, in which case the solver indicates that the desired error tolerance will not be satisfied and terminates the integration by fixing the step size to $h$ and accepting each following step without performing any step size updates anymore.

\paragraph{Initial Step-Size}
The initial step size $h_{0}$ is obtained differently in the training and testing setting. 
In training, a proposal initial step size $\tilde{h}_0$ is calculated using
\begin{equation}\label{eq:proposal-init}
    \tilde{h}_0 = \begin{cases}
\frac{\lVert \bm{z}_0 \rVert_1}{100 * \lVert \bm{g}_{\theta}\left(0, \bm{z}_0 \right)\rVert_1}, \quad \text { if } \lVert \bm{g}_0 \rVert_1 \geq 10^{-5} * \gamma \text{ and }  \lVert \bm{g}_{\theta}\left(0, \bm{z}_0 \right) \rVert_1 \geq 10^{-5} * \gamma,\\
10^{-5} , \quad \quad \quad \quad \text {\ \ otherwise,}
\end{cases}
\end{equation}
where $\gamma = b * \atol$ is determined by the batch size $b$ and the absolute error tolerance $\atol$.
Afterward, a solver step is performed using the proposal step size $\tilde{h}_0$, and the step size update rule of standard adaptive step size solvers is used in order to produce the initial step size $h_0$. 
Note that by applying the standard update rule, the solver starts the integration process with a step size for which a step acceptance is expected.
Moreover, during training the solver keeps track of an exponentially weighted average $\eta$ of the initial step sizes, where it is updated using momentum factor $\beta$, i.e. $\eta \gets (1 - \beta) \eta + \beta * h_0$.

During testing, the current $\eta$ is set as the initial step size, i.e. $h_0 = \eta$.
Observe, that in \NN verification the division in \cref{eq:proposal-init} is avoided, for which there exists only loose abstract transformations in the \deeppoly abstract domain.
Therefore, the proposed initial step size scheme decreases the approximation error in the \deeppoly abstract domain at the cost of storing and keeping track of $\eta$. 

\subsection{\solver Comparison} \label{app:adaptive_solvers_comparison}
In \cref{fig:CAS_time} we compare the reachable states, e.g. $(t, \ h)$-pairs, of the unmodifiied dopri5 \citep{dormand1980family} adaptive solver (AS) and the dopri5-based \solver (as described in the previous paragraph) after at most two steps.
In order to simplfy the computation of the reachable states, we have assumed that $\delta_{(t, \ h)} \in \left[2^{-6},2^2\right] \ \forall t,h$.

In \cref{fig:CAS_error} we compare the dopri5 AS and dopri5-based \solver with eleven different absolute error tolerances $\tau \in \{10^{-6},4.7\cdot 10^{-6}, 2.2\cdot 10^{-5},10^{-4},5 \cdot 10^{-4},2.3\cdot 10^{-3},0.01, 0.05,0.24,1,2.42\}$ on the one-dimensional nonlinear \ode $\nabla_tz = z \cdot cos \left(0.8 \cdot cos(t)^2 + t \right)$.
For each absolute error tolerance value, we sample 2000 initial states $z(0) \sim \mathcal{U}(-2.5,2.5)$ (continuous uniform distribution) and solve the resulting IVP until $T=5$, where we report the average number of performed solver steps and the absolute error of the solver.
The absolute error is calculated via $|z(5) -z_{d8}(5)|$, where $z(5)$ is the solution of either the considered AS or \solver and $z_{d8}(5)$ is the solution of the high-order adaptive solver dopri8 with absolute error tolerance $\tau_{d8} = 10^{-7}$. 

\ccc{In \cref{fig:CAS_vs_AS_on_NODEs} we compare \solver and AS solvers with respect to their absolute errors depending on the number of performed solver steps for higher-dimensional, \nodes trained on the \mnist and \fmnist datasets, using standard training with the dopri5 AS solver as described in \cref{sec:classif-details}. 
We compare dopri5-based \solver with absolute error tolerance $\tau = 0.005$ and a dopri5 AS$_\beta$ with absolute error tolerance $\tau_\beta = \tau \cdot \beta$ and compute a `ground truth' solution as reference for error computation using an AS with a 100-times smaller error tolerance, i.e. $\beta = 0.01$.
We report the mean and standard deviation of the resulting absolute error $|z(1) -z_{GT}(1)|$ as a function of the number of solver steps over the first 1000 test set samples. 

Using the same error tolerance for \solver and AS solvers, i.e. $\beta=1$, we observe for both datasets, that while \solver solvers tend to perform more solver steps than AS$_{1}$, they have significantly smaller absolute errors at the same number of solver steps.
We track this back to the conservative step-size update rule of \solver solvers. When decreasing the absolute error tolerance of the AS by factor 2, i.e. $\beta=0.5$, we observe that the AS solver tends to performs more solver steps while still yielding larger absolute errors (see \cref{fig:CAS_vs_AS_on_FMNIST_0.5}).
We thus conclude that \solver solvers are generally competitive with AS solvers.}

\begin{figure}[h]
	\centering
	\vspace{-1mm}
	\begin{subfigure}{0.28\textwidth}
		\centering
		\includegraphics[width = \textwidth]{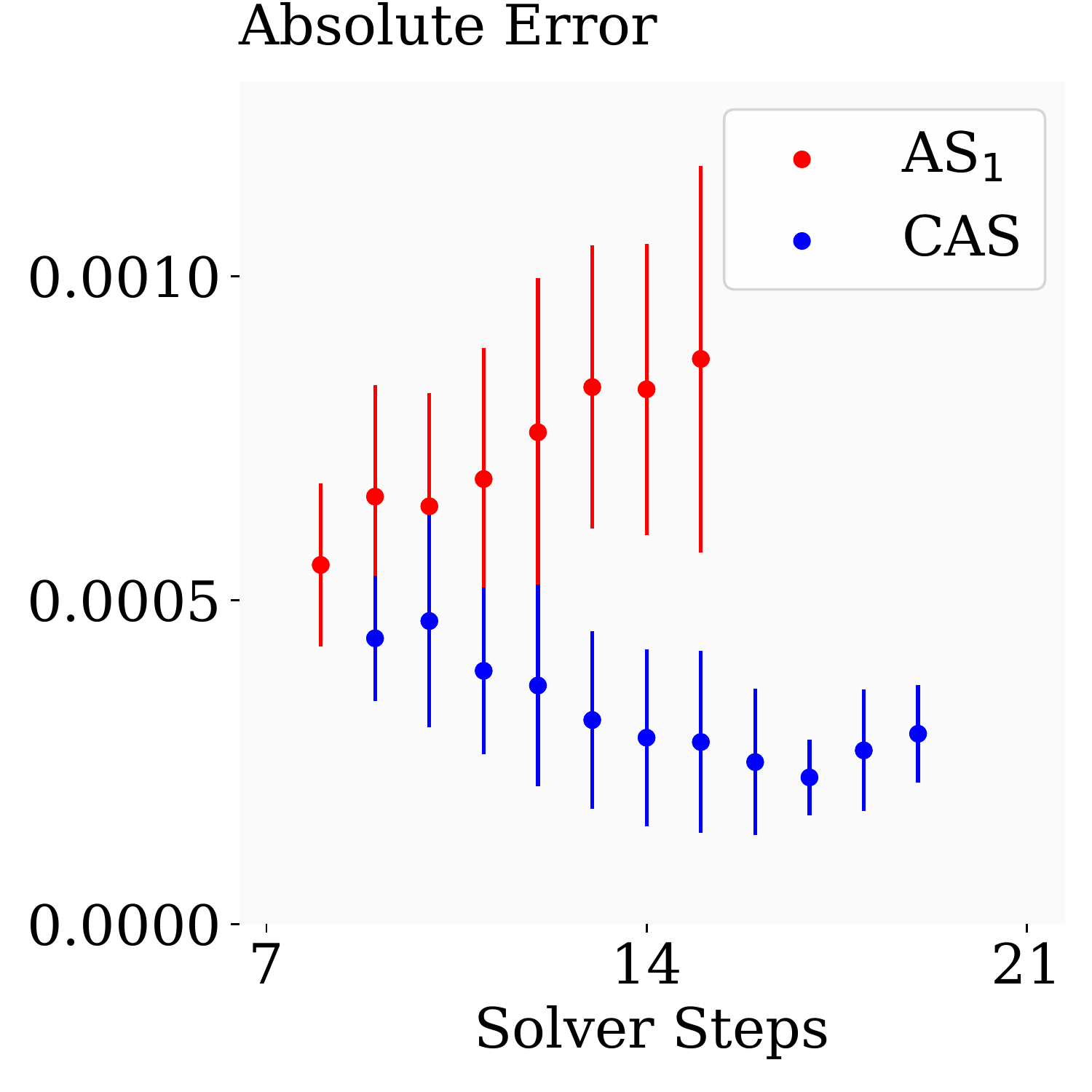} 
        \caption{\mnist with $\beta = 1$.} 
        \label{fig:CAS_vs_AS_on_MNIST}
	\end{subfigure}
	\begin{subfigure}{0.28\textwidth}
		\vspace{-1mm}
		\centering
		\includegraphics[width = \textwidth]{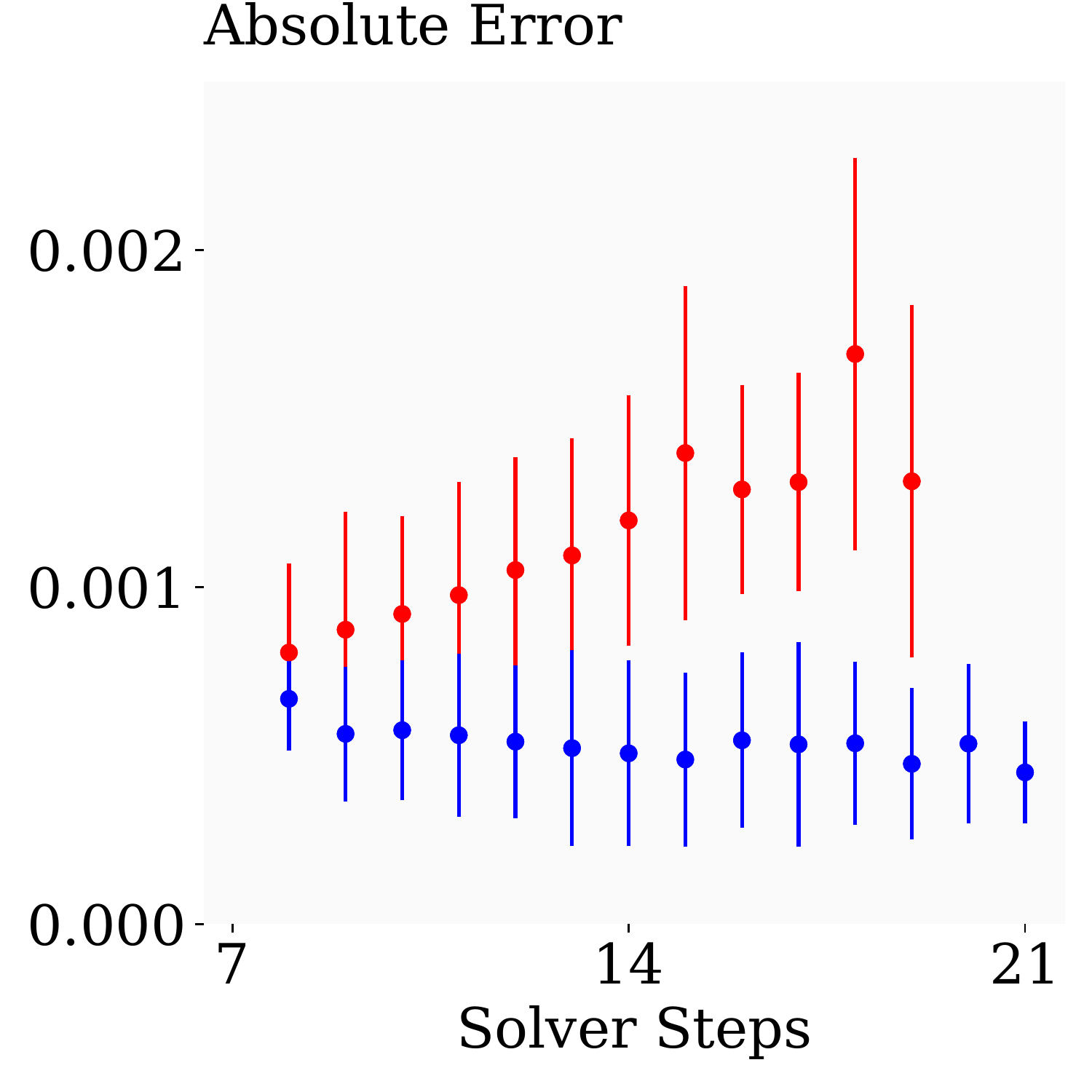}  
        \caption{\fmnist with $\beta = 1$.}
        \label{fig:CAS_vs_AS_on_FMNIST}
    \end{subfigure}
    \begin{subfigure}{0.28\textwidth}
		\vspace{-1mm}
		\centering
		\includegraphics[width = \textwidth]{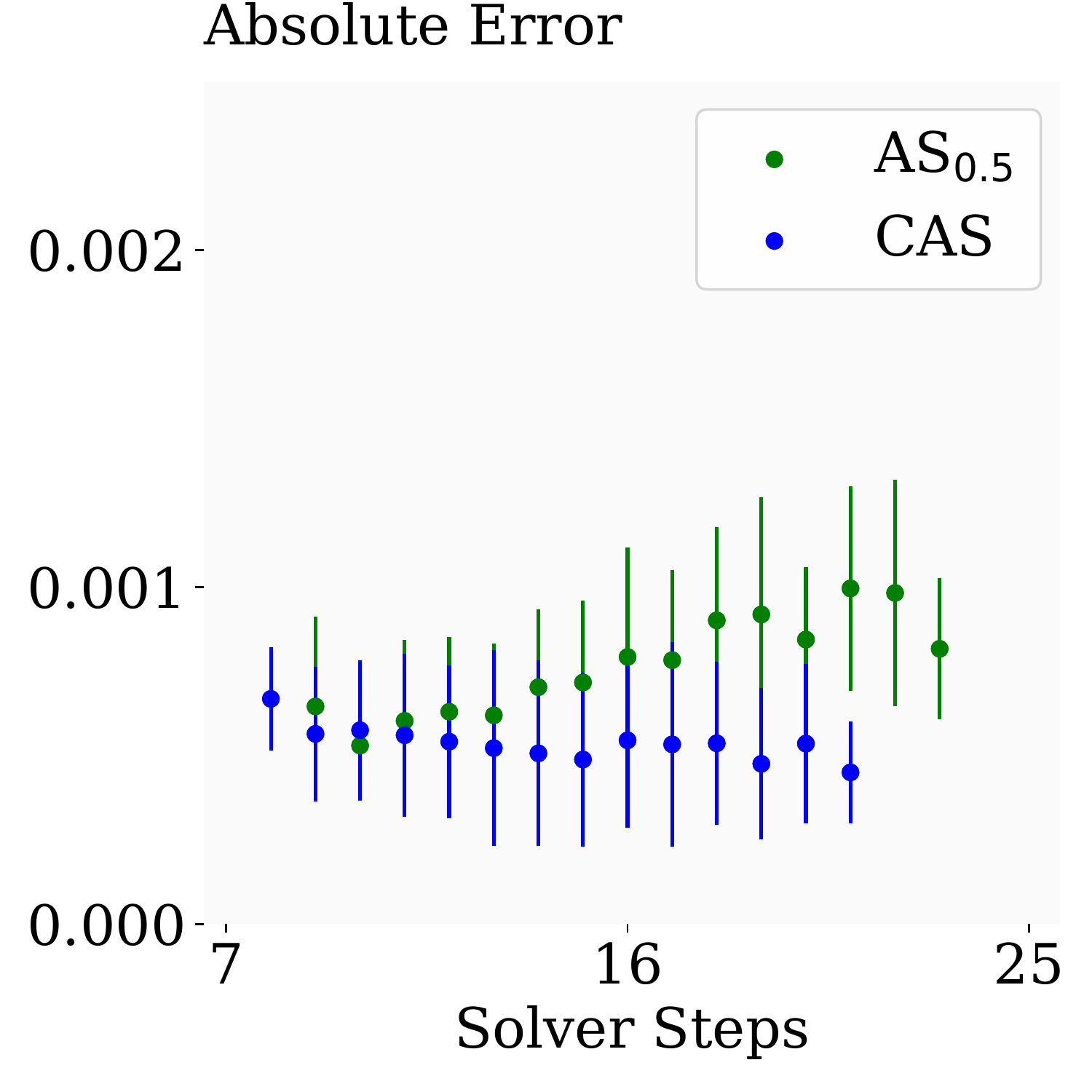}  
        \caption{\fmnist with $\beta = 0.5$.}
        \label{fig:CAS_vs_AS_on_FMNIST_0.5}
    \end{subfigure}
	\caption{Comparison of CAS and AS solvers on learned NODEs.}
	\label{fig:CAS_vs_AS_on_NODEs}
\end{figure}

\ccc{\subsection{Bound Calculation} \label{app:verification_methods}
We introduce three different approaches to compute the bounds of a neuron, namely \tool, \tool-\boxd, and \tool-Linear. 
\tool-\boxd computes the bounds by only considering interval bound propagation techniques, whereas \tool-Linear uses linear bound propagation methods (as described in \cref{sec:verification}). However, observe that when using the ReLU activation function, the selection of the slope $\lambda$ of the lower bounding function (see \cref{fig:deeppoly_relu}) allows some design choice, because each $\lambda \in [0,1]$ is valid \citep{singh2019abstract}. 
\tool-Linear selects $\lambda$ such that the area between the upper and lower bound is minimized, i.e. $\lambda = 1$ if $u_x \geq -l_x$ and $\lambda = 0$ otherwise. 
Finally, \tool is a combination of \tool-\boxd and \tool-Linear, where we compute the bounds for each neuron using both methods and use the tightest bounds to proceed. In order to further tighten the bounds, \tool additionally instantiates \tool-Linear with $\lambda = 0$ for each ReLU and \tool-Linear with $\lambda = 1$ for each ReLU.
}

\section{Classification Experiments}\label{app:classification-details}
In this section, we extend the experimental details from \cref{app:details-experiments} with emphasize on the classification experiments on the \mnist and \fmnist datasets.

\paragraph{Preprocessing} 
We have rescaled the data in both datasets such that the values are in $[0,1]$. 
Afterwards, we have standardized the data using $\mu = 0.1307 , \ \sigma =0.3081$ on the \mnist dataset and $\mu = 0.286 , \ \sigma =0.353$ on the \fmnist dataset, e.g. for input $x$ we have $x \gets \frac{x - \mu}{\sigma}$.

\paragraph{Neural Network Architecture} \label{app:classif-model}
In \cref{tab:architecture-classification}, the neural network architecture we use in classification is shown. 
The four arguments of the Conv2d layer in \cref{tab:architecture-classification} represent the input channel, output channel, kernel size, and the stride.
The two arguments of the Linear layer represents the input dimension and the output dimension.
The \node layer has $T_{end} = 1$ and \ode dynamics $\vg_{\theta}$.
Moreover, the ConcatConv2d layer takes as input a state $x$ and time $t$, where it concatenates $t$ along the channel dimension of $x$ before applying a standard Conv2d layer.
The five arguments of the ConcatConv2d layer represent the input channel, output channel, kernel size, stride and the padding.

\begin{table}[h]
\caption{The neural network architecture used in classification on the \mnist and \fmnist datasets.}
  \label{tab:architecture-classification}
\centering
\begin{tabular}{c}
\toprule
\multicolumn{1}{l}{Classification neural network $\vf_{\bs{\theta}}$ } \\
\midrule                                                       
Conv2d(1, 32, 5, 2) +  \relu                                              \\
Conv2d(32, 32, 5, 2)  +  \relu                                                   \\
\node($\vg_{\theta}$, 1)                                              \\
AdaptiveAvgPool2d                                                          \\
Linear(32,10)                                                         
\\
\bottomrule
\toprule
\multicolumn{1}{l}{\ode dynamics $\vg_{\theta}$}  \\
\midrule
$[$ConcatConv2d(33, 32, 3, 1, 1) +  \relu$]$ x2                                           \\
\bottomrule
\end{tabular}
\end{table}

\paragraph{Training Details} \label{sec:classif-details}
We used the ADAM~\citep{kingma2014adam} optimizer with learning rate 1e-3 and weight decay 1e-4 as well as batch size $b = 512$
and all the training samples in training and we have used $\mathcal{L}_{std} = \mathcal{L}_{CE}$ in \cref{eq:cert-loss}.

In provable training, we have used a warm-up training session, in which we have trained the model for 50 epochs using the fixed step size \ode solver euler with $h = \frac{1}{2}$.
Moreover, in the warm-up training session, we used the scheduler Smooth($\frac{1}{255}$, 10, 40) for the annealing of the perturbation $\epsilon$. \\
Afterward, in the actual training session, the \node layer uses a \solver as described in \cref{app:details-experiments}.
Furthermore, we train for 100 epochs using the Smooth($\epsilon_t$, 0, 60) schedule with $\epsilon_t \in \{ 0.11,0.22\}$ on the \mnist dataset and $\epsilon_t \in \{ 0.11,0.16\}$ on the \fmnist dataset.
The approximation of the abstract transformer of the \node layer uses $\kappa = 2$ in epochs 1 until 25, $\kappa = 8$ in epochs 51 until 65 and $\kappa = 4$ in all the other epochs. 
Moreover, we set $q_1 = q_2$ and use the annealing process $\text{Sin}(0.15,0.33,10, 80)$ in order to increase the value of $q_1$.
The neural network is trained using the loss function defined in \cref{eq:cert-loss} with $\omega_1 = \frac{2}{3}$ and $\omega_2 = 0.01$.

In the standard training baseline, we have trained the neural network for 100 epochs using the loss function defined in \cref{eq:cert-loss} with $\omega_1 = \omega_2 = 0$.

In the adversarial training baseline we have trained the neural network for 100 epochs, where the samples from the dataset are attacked using PGD$(\epsilon,N= 10,\alpha = \frac{\epsilon}{5}, \mathcal{L}_{CE})$ prior to being fed into the model as input.
Moreover, we use Smooth($\epsilon_t$, 5, 65) for the annealing of $\epsilon$ and $\epsilon_t = 0.11$ on both datasets.
We use the loss function in \cref{eq:cert-loss} with $\omega_1 = \omega_2 = 0$ in training.

Furthermore, we want to emphasize that whenever we are considering abstract input regions, e.g. in  provable training and adversarial training, we do not allow perturbations outside of the [0,1] interval.

\paragraph{Evaluation Details}
In order to obtain the adversarial accuracies reported in \cref{tab:results_cls}, we have used the PGD$(\epsilon,N= 200,\alpha = \frac{1}{40}, \mathcal{L}_{CE})$ attack with $\epsilon \in \{ 0.1,0.15, 0.2\}$ on the \mnist dataset and $\epsilon \in \{ 0.1,0.15\}$ on the \fmnist dataset.

\section{Further Details for Time-Series Forecasting Experiments}\label{app:time-series-forecasting-details}
In this section, we extend the experimental details from \cref{app:details-experiments} with emphasize on the time-series forecasting task on the \physio dataset.
Moreover, we have made use of the code provided by \citet{rubanova2019latent}\footnote{\hyperlink{https://github.com/YuliaRubanova/latent\_ode}{https://github.com/YuliaRubanova/latent\_ode}} for the fetching of the dataset and parts of the latent \ode architecture.

\paragraph{\physio Preprocessing}\label{app:physionet-preprocessing}
The \physio dataset contains data from the first 48 hours of a patients stay in intensive care unit (ICU). 
The dataset consists of 41 possible features per observed measurement, where the measurements are made at irregular times and not all possible features are measured.
We round up the time steps to three minutes, which results in the length of the time-series being at most $48 \cdot 20 + 1 = 961$.

Moreover, we remove four time-invariant features and additionally two categorical features from the series, namely the Gender, Age, Height, ICUType, GCS, and MechVent.
The removed features are inserted in an initial state $\vx_{0} \in \mathbb{R}^6$ of the time-series, which is used to initialize the hidden state of the encoder.
Note that there is exactly one measurement for the features Gender, Age, Height, and ICUType, which we used unaltered as the first four entries of the initial state $\vx_{0}$.
On the other hand, in the case where we want to predict a value in the future while only using the first $L'$ entries of an input series, there can be multiple or no measurements for the GCS and MechVent features among the first $L'$ entries of the series.
If there are measurements made for the GCS feature, we use the average of the observed values as the fifth entry of $\vx_{0}$, whereas if there are measurements for the MechVent feature we set the sixth entry of $\vx_{0}$ to 1.
Otherwise, if there are no measurements for the two aforementioned features their corresponding entry in $\vx_{0}$ is set to zero.

Additionally, we clip the measurements for features with high noise or atypical values.
Concretely, we clip the Temp feature to the [32,45] interval, the Urine feature to the [0,2000] interval, the WBC feature to the [0,60] interval, and the pH feature to the [0,14] interval.

Furthermore, we split the dataset into a training set containing 7200 time-series, validation set containing 400 time-series, and testing set containing 400 time-series.

We normalize the features to be normally distributed, where we estimate the mean and standard deviation of each feature using the training set.
The normalization is used for all features except the categorical features (Gender, ICUType, GCS, MechVent) and the features Fi02 and Sa02, which represent a ratio. The categorical features are used unaltered, whereas the ratios are rescaled in order to be in the [0,1] interval.

Finally, we introduce three different data modes \emph{6h}, \emph{12h} and \emph{24h}, which we consider for the time-series forecasting task.
The data modes differ in the number of entries $L'$ which are used as input in order to estimate the final data point of a series.
When considering the time-series $\vx^L_{ts} = \{(\vx_{(i)},t_{(i)})\}_{i=1}^L$ and the data mode \emph{6h}, the number of entries used as input is $L'_6 = \max_{i \in [L]} i$ such that $t_{(i)}\leq t_{(L)}-6$, i.e. we try to predict at least six hours into the future.
The data modes \emph{12h} and \emph{24h} are defined in the same way, where we try to predict at least 12 or 24 hours into the future.
Furthermore, for a fixed time-series it follows that $L'_{6}\geq L'_{12} \geq L'_{24}$.

\paragraph{Time-Series Forecasting Architecture} \label{app:forecasting-model}
In \Cref{tab:architecture-forecasting}, we show the main components of the latent \ode architecture, which we use for the time-series forecasting task on the \physio dataset. 
In the \node layer of the encoder $\ve_{\bs{\theta}}$ we use a one-step euler \ode solver, where the step size $h$ depends on the measured time points in the input time-series.
On the other hand, the \node layer in the decoder $\vd_{\bs{\theta}}$ uses the \solver as specified in \cref{app:details-experiments} and the final integration time depends on the time-series point we want to estimate, e.g. if we estimate $\vx_{(L)}$ we use  $T_{end} = t_{(L)}$.

\begin{table}[h]
\caption{The main components of the latent \ode architecture used in time-series forecasting on the \physio dataset.}
  \label{tab:architecture-forecasting}
\centering
\begin{tabular}{cc}
\toprule
\multicolumn{2}{l}{Encoder $\ve_{\bs{\theta}}$}                           \\
\midrule
\multicolumn{2}{c}{Linear(6,80) + \relu}                    \\
\multicolumn{2}{c}{GRU-Unit $\bm{f}^{\text{GRU}}_{\bs{\theta}}$}                          \\
\multicolumn{2}{c}{Linear(80,100)  + \relu}                 \\
\multicolumn{2}{c}{Linear(100,40)}                               \\
\bottomrule
\toprule
\multicolumn{2}{l}{GRU-Unit $\bm{f}^{\text{GRU}}_{\bs{\theta}}$}                          \\
\midrule
\multicolumn{1}{l}{$f_{z}$}        & \node($\bm{g}^{\ve}_{\bs{\theta}}$) \\
\midrule
\multicolumn{1}{l}{$f_{u}, f_{r}$} & Linear(115,50) + \relu \\
\multicolumn{1}{l}{}               & \ \ Linear(50,40) + Sigmoid     \\
\midrule
\multicolumn{1}{l}{$f_{s} $}       & Linear(115,50) + \relu \\
\multicolumn{1}{l}{}               & Linear(50,80)               \\
\bottomrule
\toprule
\multicolumn{2}{l}{\ode dynamics $\bm{g}^{\ve}_{\bs{\theta}}$}                \\
\midrule
\multicolumn{2}{c}{$[$Linear(40,40) + \relu $]$ x3}         \\
\multicolumn{2}{c}{Linear(40,40)}                                \\
\bottomrule
\toprule
\multicolumn{2}{l}{Decoder $\vd_{\bs{\theta}}$}                           \\
\midrule
\multicolumn{2}{c}{NODE($\bm{g}^{\vd}_{\bs{\theta}}$)}                             \\
\multicolumn{2}{c}{Linear(20,35)}                                \\
\bottomrule
\toprule
\multicolumn{2}{l}{\ode dynamics $\bm{g}^{\vd}_{\bs{\theta}}$}                \\
\midrule
\multicolumn{2}{c}{Linear(20,40) + \relu}                   \\
\multicolumn{2}{c}{$[$Linear(40,40) + \relu $]$ x2}         \\
\multicolumn{2}{c}{Linear(40,20)}                             \\
\bottomrule
\end{tabular}
\end{table}

\paragraph{Training Details} \label{sec:forecasting-details}
We have used batch size $b = 128$ and $\mathcal{L}_{std} = \mathcal{L}_{f}$ in \cref{eq:cert-loss} with $\mathcal{L}_{f}$ defined in \cref{eq:forecasting-loss-background} and $\gamma = 30000$.
Moreover, we assume that the initial state of the generative model of the time series has prior distribution $\mathcal{N}(0,1)$. 
What is more, since not all feature values are observed in each measurement, we want to emphasize that only the observed features are used to evaluate any metric.
For example, if the final data point $\vx_{L}$ has measured features at the entries in the set $M \subseteq [35]$ and we obtain the estimate $\hat{\vx}_{L}$, the MAE is given by
\begin{align}
    \text{\mae}(\vx_{L},\hat{\vx}_{L}) = \frac{1}{|M|}\sum_{j \in M} |x_{L,j}-\hat{x}_{L,j}|.
\end{align}
Additionally, as our validation metric, we use the MAE with concrete inputs in all experiments in order to evaluate the performance of the model on the validation set.
We have trained the models on the random seeds 100, 101, and 102\footnote{Some models were trained with seed 103.}.

Moreover, observe that in a batched input setting the sequence length of the individual time-series can be different, and also the time in which measurements are made differs. 
In order to circumvent this issue and allow batched training, we take the union of the time points and extend each individual series to contain all time points observed in the batch, where we add data points with no measured features to each series.
Furthermore, in batched training, the GRU-unit of latent \ode only performs an update to the hidden state to those inputs in the batch, for which at least one feature was observed in the data point at the currently considered time.

In standard training, we have trained the latent \ode for at most 120 epochs, where after each epoch we evaluate the performance of the model on the validation set and use the model with the best performance on the validation set in testing. 
Note, that if the performance on the validation does not improve for 10 epochs we apply early stopping.
Furthermore, ADAM~\citep{kingma2014adam} was used as optimizer with learning rate 1e-3 and weight decay 1e-4 and we have used $\omega_1 = \omega_2 = 0$ in \cref{eq:cert-loss}. 

In provable training, we have trained the latent \ode for 120 epochs, where we have used the scheduler Smooth($\epsilon_t$, 5, 65) for the perturbation with $\epsilon_t \in \{0.1,0.2\}$.
The approximation of the abstract transformer of the \node layer in the decoder $\vd_{\bs{\theta}}$ uses $\kappa = 1$ in all epochs, whereas the \node layer in the encoder $\ve_{\bs{\theta}}$ has due to the chosen \ode solver always only one possible trajectory.
Moreover, in the \node layer of $\vd_{\bs{\theta}}$, we set $q_1 = q_2$ and use the annealing process $\text{Sin}(0.15,0.33,10, 80)$ in order to increase the value of $q_1$.
Furthermore, the abstract ratio $\rho$ is initialized as $\rho = 0.1$ and we increase its value by 0.05 at the end of epochs $\{10,15\}$ and by 0.1 at the end of epochs $\{10 + 5\cdot i\}_{i=2}^{9}$.
Moreover, ADAM was used as optimizer with learning rate 1e-3 and weight decay 1.
Furthermore, as soon as the target perturbation is reached ($\epsilon' = \epsilon_t$), we evaluate the performance of the model on the validation set after each epoch and use the model with the best performance on the validation set in verification. 

\paragraph{Evaluation Details}
In order to obtain the adversarial accuracies reported in \cref{tab:results_time}, we have used the PGD$(\epsilon,N= 200,\alpha = \frac{1}{40}, \text{MAE})$ attack with $\epsilon \in \{ 0.05,0.1, 0.2\}$ on all data modes of the \physio dataset.

\section{Trajectory Attacks} \label{app:trajectory-attack}
In order to describe the used attacking procedure, let us denote by $\delta_1$ the local error estimate of the solver in the first step, e.g. $\delta_1 = \delta_{(0,\ h_0)}$, and by $\delta_2$ the local error estimate from the second step. Moreover, assume that we use a \solver with update factor $\alpha$.

We describe the attack for a single $\delta_{i}$ with $i = 1,2$ first and afterward how to combine them.
The loss function $\mathcal{L}_{att}(\vz_0)$ we try to maximize during the attack, depends on the value of $\delta_i$, where in the case that $\delta_{i} \in [0,\tau_{\alpha}] \cup [\frac{\tau_{\alpha}+ 1}{2},1]$, we have $\mathcal{L}_{i}(\vz_0) = \delta_{i}$, whereas otherwise $\mathcal{L}_{i}(\vz_0) = -\delta_{i}$ is used.
Hence, we try to decrease or increase the error estimate $\delta_i$ depending on the closest decision boundary, such that a different update is performed.

The attacks are performed by using the $\{\text{PGD}(\epsilon,100,\frac{1}{40},\mathcal{L}_{att,m})\}_{i=-1}^{5}$ attacks with $\epsilon \in \{0.1,0.15,0.2\}$ and we define $\mathcal{L}_{att,m}$ next.
The parameter $m$ specifies how to combine the loss functions for the individual local error estimates $\delta_1$ and $\delta_2$, where for $m = -1$ we use $\mathcal{L}_{att,-1}(\vz_0) = \mathcal{L}_{1}(\vz_0)$, for $m = 0$ we use $\mathcal{L}_{att,0}(\vz_0) = \mathcal{L}_{1}(\vz_0) + \mathcal{L}_{2}(\vz_0) $ and for $m \geq 1$ we use in PGD iteration $j$ the loss $\mathcal{L}_{att,i}(\vz_0) = \mathcal{L}_{2}(\vz_0)$ if $j \text{ mod } m = 0$ and otherwise $\mathcal{L}_{att,i}(\vz_0) = \mathcal{L}_{1}(\vz_0)$.

In our experiments, we use the attacks with $-1\leq m \leq 5$ for the same input $\vz_0$ and as soon as we have successfully found $\vz_0' \in \bc{B}^{\epsilon}(\vz_0)$ such that $\Gamma(\vz_0) \neq \Gamma(\vz_0')$ holds, the attack is stopped and considered to be successful.

\section{\deeppoly Toy Dataset \& LP Baseline} \label{app:dpcap_details}
In this section, we describe the generation of the \deeppoly toy dataset and the used LP baseline in the \lcap experiments in \cref{sec:results-relu-method}.
In order to do so, we define the discrete uniform distribution $\mathcal{U}(\mathcal{X})$ over a set $X = \{\vx_i\}_{i=1}^{n}$ and the continuous uniform distribution $\mathcal{U}(a,b)$ on a bounded domain $[a,b]$, i.e. $-\infty < a < b < \infty$.
The former distribution is a categorical distribution with $p_i = \frac{1}{n}$ $\forall \ i \in [n]$, whereas the latter distribution has probability density function $p_{\mathcal{U}}(x) = \frac{1}{b-a} \ \forall x' \in [a,b]$ and $p_{\mathcal{U}}(x) = 0$ otherwise.

\paragraph{\lcap Toy Dataset } \label{sec:DP-dataset}
To generate $m$ different linear constraints in order to describe a random relation between activation $y\in \mathbb{R}$ and activations $\vx\in \mathbb{R}^d$.
We only describe the process for the upper bounds of the linear constraints, since the construction of the lower bounding constraint follows analogously.
Additionally, we define the cosine similarity between two vectors as $\csim(\bm{a},\bm{b}) = \frac{\sum_{i=1}^{d} a_i \cdot b_i}{\lVert \bm{a} \rVert_2 \lVert \bm{b} \rVert_2}$ with $\lVert \bm{a} \rVert_2 = \left(\sum_{i=1}^{d}a_i^2\right) ^{\frac{1}{2}}$.
We ensure that the average cosine similarity among the produced upper bounds is within $[0.975, \ 0.99]$.
The lower bound on the similarity is included since we assume that all linear constraints describe the same relation and therefore we expect them to be similar.
On the other hand, the upper bound on the similarity is included such that there are at least some differences between the constraints and the \lcap is harder to solve.

Furthermore, we define the functions $g_1(d) = 5 \cdot \left(\min\left(1,\frac{20}{d +1}\right)\right)^2 $, $g_2(d) = \beta \cdot \min\left(1,\frac{5}{d +1} \cdot \left\lceil \frac{d+1}{50}\right\rceil\right)$ with $\beta = 3$ and the ceiling function $\lceil z \rceil = \min\{ n\in \mathbb{N} | n\geq z\}$, and $g_{\bm{\alpha}}(\vx) =  \sum_{j = 1}^d \alpha_j\cdot x_j + \alpha_{d+1}$ for any $\bm{\alpha} \in \mathbb{R}^{d+1}$.

First, we construct the abstract input domain $\mathcal{X}$, where for each entry $x_j$ we sample $z_1,z_2 \sim \mathcal{U}(-g_1(d),g_1(d))$ and set $l_{x_j} = \min(z_1,z_2)$ and $u_{x_j} = \max(z_1,z_2)$.

Afterwards, we sample the coefficients $a_{j}\sim \mathcal{U}\left(-\frac{\beta}{2},\frac{\beta}{2}\right) \ \forall j \in [d+1]$  and fix the relation between $\vx$ and $y$ as $y = g_{\bm{a}}(\vx)$. 
Next, we sample the coefficients $w_{j}^0\sim \mathcal{U}(-\beta,\beta) \ \forall j \in [d+1]$ and define the proposal upper bound $g_{\bm{w^0}}(\vx)$. We apply an upper bounding update to the bias term if it is not a proper upper bound, i.e. $w_{d+1}^0 \gets w_{d+1}^0 - \min_{\vx' \in \mathcal{X}} g_{\bm{w^0-a}}(\vx')$ if $\min_{\vx' \in \mathcal{X}} g_{\bm{w^0-a}}(\vx') < 0$.
The proposal upper bound is accepted as the upper bound if $|w_{d+1}^0| \leq 2\cdot \beta$ and otherwise we repeat the procedure until we have an accepted upper bound.

Afterward, we initialize the upper bounding set $U = \{ \}$, which is iteratively augmented until its cardinality is $m$.
In the first iteration we sample $\Delta_{j}^1\sim \mathcal{U}(-g_2(d),g_2(d)) \ \forall j \in [d+1]$ and define $\bm{w^{1}} = \bm{w^0} + \Delta^1$.
Moreover, the bias term of $g_{\bm{w^{1}}}$ is corrected using the upper bounding update, such that we have $g_{\bm{w^{1}}}(\vx') \geq g_{\bm{a}}(\vx') \ \forall \vx' \in \mathcal{X}$. We include $\bm{w^{1}}$ to $U$ if $|w_{d+1}^1| \leq 2\cdot \beta$, and otherwise repeat until the iteration is accepted. \newline
In the $i$-th iteration, $\bm{w^{i}}$ is obtained by applying the same procedure as in the first iteration. 
However, $\bm{w^{i}}$ is only included to $U$ if $|w_{d+1}^i| \leq 2\cdot \beta$ and $\frac{1}{|U|}\sum_{k = 1}^{|U|}\csim(\bm{w^{i}},\bm{w^{k}}) \geq 0.975$, otherwise we repeat the calculation of $\bm{w^{i}}$.

As soon as the cardinality of $U$ equals $m$, we calculate the average similarity of the vectors in $U$ and accept the set $U$ if the similarity is less than 0.99, i.e. $\frac{1}{(m-1) \cdot (m-2)}\sum_{i=1}^m\sum_{k = i+ 1}^m \csim(\bm{w^{i}},\bm{w^{k}})\leq 0.99$. 
Otherwise, the set is rejected and we reinitialize the process from the beginning.
If the set is accepted, we define the linear upper bounding constraints using $u^{i} = g_{\bm{w^{i}}}$ for $i \in [m]$.

Observe that the generation process is probabilistic and we often reject proposal coefficients and sets.
Hence, in order to avoid a non-terminating process, we limit the number of sampled vectors to 35000.

\paragraph{LP Baseline} \label{sec:dpcap-baselines}
We have used LP(8, 50, 40) as a baseline for the \lcap toy dataset experiment, where for a \lcap with $m$ different constraints that describe the relation between $z\in \mathcal{Z}\subseteq \mathbb{R}^d$ and $y \in \mathbb{R}$ the baseline works as follows.
The LP baseline initially defines the set $\mathcal{Z}' = \{\vz'_{(j)}\}_{j=1}^{8 \cdot d}$ with $\vz'_{(j)} \sim \mathcal{U}(\partial\mathcal{Z}) \ \forall j \in [8\cdot d]$, where $\partial\mathcal{Z}$ are the corners of $\mathcal{Z}$, and solves the resulting optimization problem when replacing $\mathcal{Z}$ with $\mathcal{Z}'$ in \cref{eq:LCAP}. 
We denote the optimal solution of the simplified optimization problem by $u^{\mathcal{Z}'}$, which is obtained by using a commercial linear program solver (GUROBI~\citep{gurobi}).
Note that due to the linear form of all the constraints, it is enough to only consider the $2^d$ points in $\partial\mathcal{Z}$ in the optimization constraint of \cref{eq:LCAP}. 

Observe that since we have loosened the restrictions, we may have that $u^{\mathcal{Z}'}$ is unsound in $\partial\mathcal{Z}$, i.e. it exists some $\vz' \in \partial\mathcal{Z}$ and $i \in [m]$ such that $u^{\mathcal{Z}'}(\vz') < u^{i}(\vz')$. \newline
If $u^{\mathcal{Z}'}$ is sound it is used as the solution of the LP baseline, otherwise for all $i\in [m]$ that violate the soundness check, we add $\hat{\vz}^{i} = \argmin_{\vz' \in \partial\mathcal{Z}}u^{\mathcal{Z}'}(\vz') - u^{i}(\vz')$ to the current $\mathcal{Z}'$.
Moreover, for each $\hat{\vz}^{i}$ we produce the corner points $\{\vz^{i,k}\}_{k=1}^{40 - 1}$ and add them to $\mathcal{Z}'$ as well, where we have $z^{i,k}_j = \hat{z}^{i}_j$ with probability 0.75 and else $z^{i,k}_j = l_{z_j} + u_{z_j} - \hat{z}^{i}_j \  \forall k \in [40 -1]$, $ \forall j \in [d]$.

This process is repeated at most 50 times and if the solution $u^{\mathcal{Z}'}$ is still unsound after 50 iterations, we add $\max_{i\in [m] , \vz' \in \partial\mathcal{Z}}u^{i}(\vz') - u^{\mathcal{Z}'}(\vz')$ as a correction bias.

\ccc{\section{Additional Experiments} \label{app:additional_exp}
\subsection{Comparison CAS and AS}\label{app:additional_exp_CAS}
To further compare \solver and AS solvers, we train and evaluate \nodes of the same architecture (see \cref{app:classification-details}) with either CAS or AS using both standard and adversarial training ($\epsilon_t = 0.11$).
We report mean and standard deviation of the resulting standard and adversarial accuracy on \mnist and \fmnist across three runs in \cref{tab:results_AS}.
We observe that while the mean performance with AS is better than that with CAS solvers in more settings than vice-versa, across both datasets and all perturbation magnitudes, there is not a single setting, where the $\pm 1$ standard deviation ranges do not overlap. Further, we observe the same trends regardless which solver we use.
We thus conclude that any performance difference between \solver and AS solvers is statistically insignificant.
\begin{table}[h]
    \centering	
    \caption{Means and standard deviations of the standard (Std.) and adversarial (Adv.) accuracy evaluated using CAS or AS on the first 1000 test set samples.\vspace{-2mm}}
    \begin{adjustbox}{width=0.9\columnwidth,center}
    \begin{threeparttable}
    	\renewcommand{\arraystretch}{0.98}
    \begin{tabular}{lccccccc}
		\toprule
        \multirow{2.5}*{Dataset} & \multirow{2.5}*{Training Method} & \multirow{2.5}*{ODE Solver} & &  \multirow{2.5}*{Std. [\%]} & \multicolumn{3}{c}{Adv. [\%]}\\
       	 \cmidrule(rl){6-8} 
       	 &&&&& $\epsilon = 0.10$  & $\epsilon = 0.15$ &  $\epsilon = 0.20$ \\
         \midrule
         
         \multirow{5}*{\mnist}& \multirow{2}*{Standard}  &  AS & & \textbf{99.2}$^{\pm 0.1}$&\textbf{24.5}$^{\pm 2.0}$&1.9$^{\pm 0.7}$& 0.0$^{\pm 0.2}$\\
        & & CAS& & 98.8$^{\pm 0.4}$&23.2$^{\pm 3.5}$&\textbf{2.5}$^{\pm 1.6}$&\textbf{0.3}$^{\pm 0.2}$ \\
         \cmidrule(rl){2-3}
                              &\multirow{2}*{Adv.} &  AS &     & \textbf{99.2}$^{\pm 0.2}$&\textbf{95.9}$^{\pm 0.2}$&\textbf{88.5}$^{\pm 0.6}$& 54.6$^{\pm 2.4}$\\
                              && CAS    & & \textbf{99.2}$^{\pm 0.1}$& 95.4$^{\pm 0.4}$&88.3$^{\pm 0.6}$& \textbf{59.4}$^{\pm 3.2}$ \\
                             
    \cmidrule(rl){1-3}

    \multirow{5}*{\fmnist}& \multirow{2}*{Standard}  &  AS & & \textbf{90.3}$^{\pm 0.4}$&\textbf{1.3}$^{\pm 1.6}$&\textbf{0.5}$^{\pm 0.7}$\\
        & & CAS& & 88.6$^{\pm1.2}$&0.1$^{\pm 0.1}$&0.0$^{\pm 0.0}$\\
         \cmidrule(rl){2-3}
                              &\multirow{2}*{Adv.} &  AS &     & 80.8$^{\pm 0.5}$&\textbf{70.3}$^{\pm 0.3}$&\textbf{53.6}$^{\pm 3.1}$\\
                              && CAS    & & \textbf{80.9}$^{\pm 0.7}$&70.2$^{\pm 0.5}$ &47.1$^{\pm 3.7}$ \\

    \bottomrule
    \end{tabular}
    \label{tab:results_AS}
    \end{threeparttable}
	\end{adjustbox}
\vspace{-4mm}
\end{table}

\subsection{Comparison \tool and TisODE}
We compare our certified training via \tool to the heuristic defence of \citet{yan2019robustness}, which introduce time-invariant steady neural ODEs (TisODEs) using a pre-trained TisODE model from \citet{yan2019robustness}\footnote{\hyperlink{https://github.com/HanshuYAN/TisODE}{https://github.com/HanshuYAN/TisODE}} with $141\,130$ trainable parameters and a \tool-trained NODE with $45\,866$ parameters.
Reporting standard and adversarial accuracies for \mnist in \cref{tab:results_tisode}, we observe that while the TisODE has a higher standard accuracy, its adversarial accuracy quickly decreases with perturbation size, falling to $55.5\%$ at $\epsilon=0.2$, where the \tool-trained \node still has $84.5\%$ adversarial accuracy. We highlight that TisODEs are not trained with future certification in mind, explaining the gap in standard accuracy.

\begin{table}[t]
    \centering	
    \caption{Comparison of \tool-trained and TisODEs \citep{yan2019robustness} with respect to standard (Std.) and adversarial (Adv.) accuracy on the first 1000 test set samples of the MNIST dataset.\vspace{-2mm}}
    \begin{adjustbox}{width=0.7\columnwidth,center}
    \begin{threeparttable}
    	\renewcommand{\arraystretch}{0.98}
    \begin{tabular}{lccccc}
		\toprule
       	 \multirow{2.5}*{Training Method} & &  \multirow{2.5}*{Std. [\%]} & \multicolumn{3}{c}{Adv. [\%]}\\
       	 \cmidrule(rl){4-6} 
       	 &&& $\epsilon = 0.10$  & $\epsilon = 0.15$ &  $\epsilon = 0.20$ \\
       	 \midrule
         TisODE \citep{yan2019robustness} & & \textbf{99.3}& \textbf{93.1}&78.6&55.5\\
         GAINS ($\epsilon_t = 0.22$)    & & 91.8&88.5&\textbf{86.8}&\textbf{84.5} \\
                            \bottomrule 
    \end{tabular}

    \label{tab:results_tisode}
    \end{threeparttable}
	\end{adjustbox}
\end{table}

\begin{table}[t]
	\centering	
		\caption{Means and standard deviations of the standard (Std.) and certified (Cert.) accuracy obtained using \tool, \tool-Linear and \tool-\boxd evaluated on the first 1000 \fmnist test set samples.\vspace{-2mm}}
	\begin{adjustbox}{width=\columnwidth,center}
		\begin{threeparttable}
			\renewcommand{\arraystretch}{0.98}
			\begin{tabular}{ccccccccc}
				\toprule
				\multirow{2.5}*{$\epsilon_t$} &  & \multirow{2.5}*{Std. [\%]} & \multicolumn{3}{c}{$\epsilon = 0.10$} & \multicolumn{3}{c}{$\epsilon = 0.15$} \\
				\cmidrule(rl){4-6} \cmidrule(rl){7-9} 
				&&& \tool-\boxd Cert. [\%] & \tool-Linear Cert. [\%]  & \tool Cert. [\%] & \tool-\boxd Cert. [\%]  & \tool-Linear Cert. [\%] &\tool Cert. [\%] \\
				\midrule
				
				0.11    &  & 75.1$^{\pm 1.2}$ &44.2$^{\pm 5.5}$&56.3$^{\pm 1.4}$& 62.5$^{\pm 1.1}$&  3.5$^{\pm 1.4}$&8.4$^{\pm 2.3}$ & 13.3$^{\pm 3.1}$ \\
				0.16    &  & 71.5$^{\pm 1.7}$ &47.0$^{\pm 5.7}$&54.7$^{\pm 2.5}$& 61.3$^{\pm 2.7}$ &36.8$^{\pm 5.2}$&42.7$^{\pm 1.4}$& 55.0$^{\pm 4.3}$  \\
				\bottomrule
			\end{tabular}
			
			\label{tab:results_IBP}
		\end{threeparttable}
	\end{adjustbox}
\end{table}

\subsection{Ablation \tool Verification}
To analyse the effect of combining linear-bound propagation with interval bound propagation, discussed in \cref{app:verification_methods}, we conduct two experiments: First, we compare the certified accuracies obtained with \tool to \tool-Linear, a version only using linear-bound propagation, and \tool-\boxd, a version only using interval bound propagation (both use our trajectory graph construction).
Second, we compare the bounds on output logit differences obtained with \tool, \tool-Linear and \tool-\boxd to those obtained via an adversarial attack using PGD.

In \cref{tab:results_IBP}, we report the certified accuracies obtained with \tool, \tool-Linear and \tool-\boxd on the \fmnist dataset and observe that \tool outperforms the other methods in every setting, showcasing that \tool inherits benefits from both linear- and interval bound propagation. Moreover, we additionally observe that using \tool-Linear results in higher accuracies than using \tool-\boxd, demonstrating the importance of linear bound propagation and thus \lcapm for our method \tool.

\begin{wrapfigure}[18]{r}{0.58\textwidth}
	\centering
	\vspace{-1mm}
	\begin{subfigure}{0.49\linewidth}
		\centering
		\includegraphics[width = 0.95\textwidth]{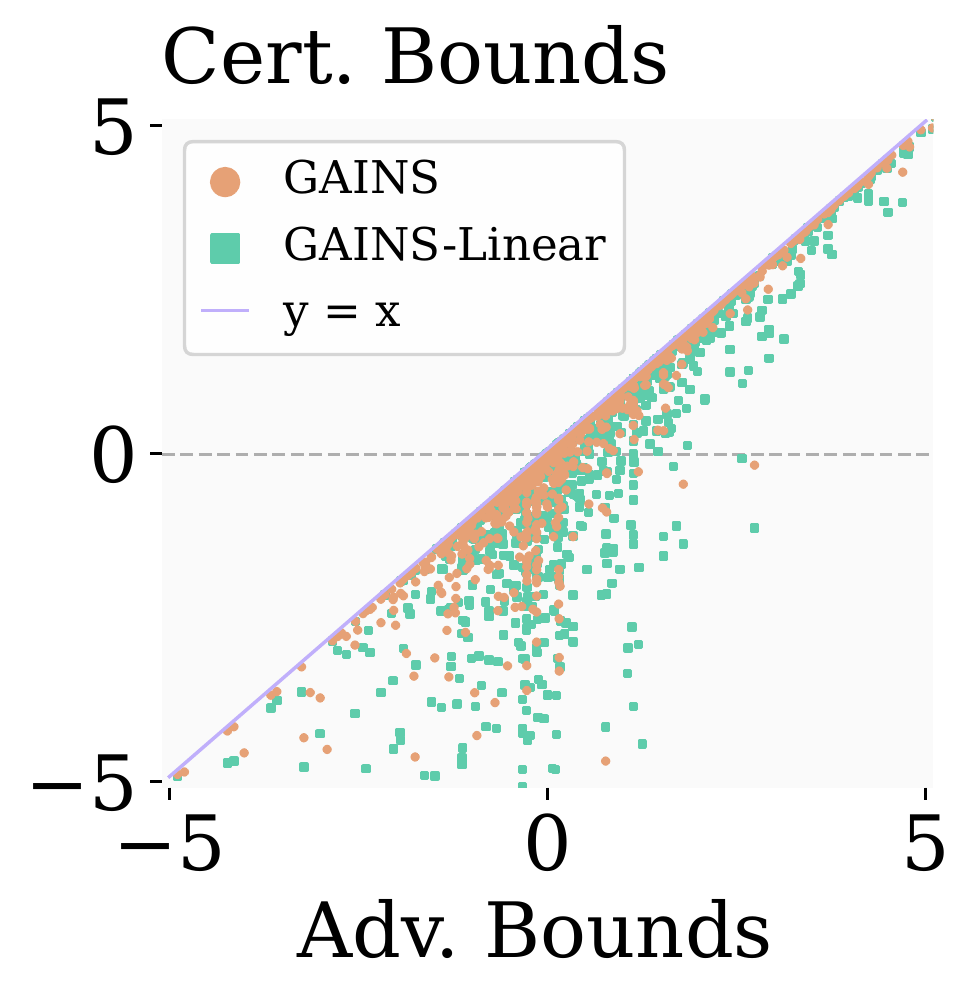} 
	\end{subfigure}
	\hfil
	\begin{subfigure}{0.49\linewidth}
		\vspace{-1mm}
		\centering
		\includegraphics[width = 0.95\textwidth]{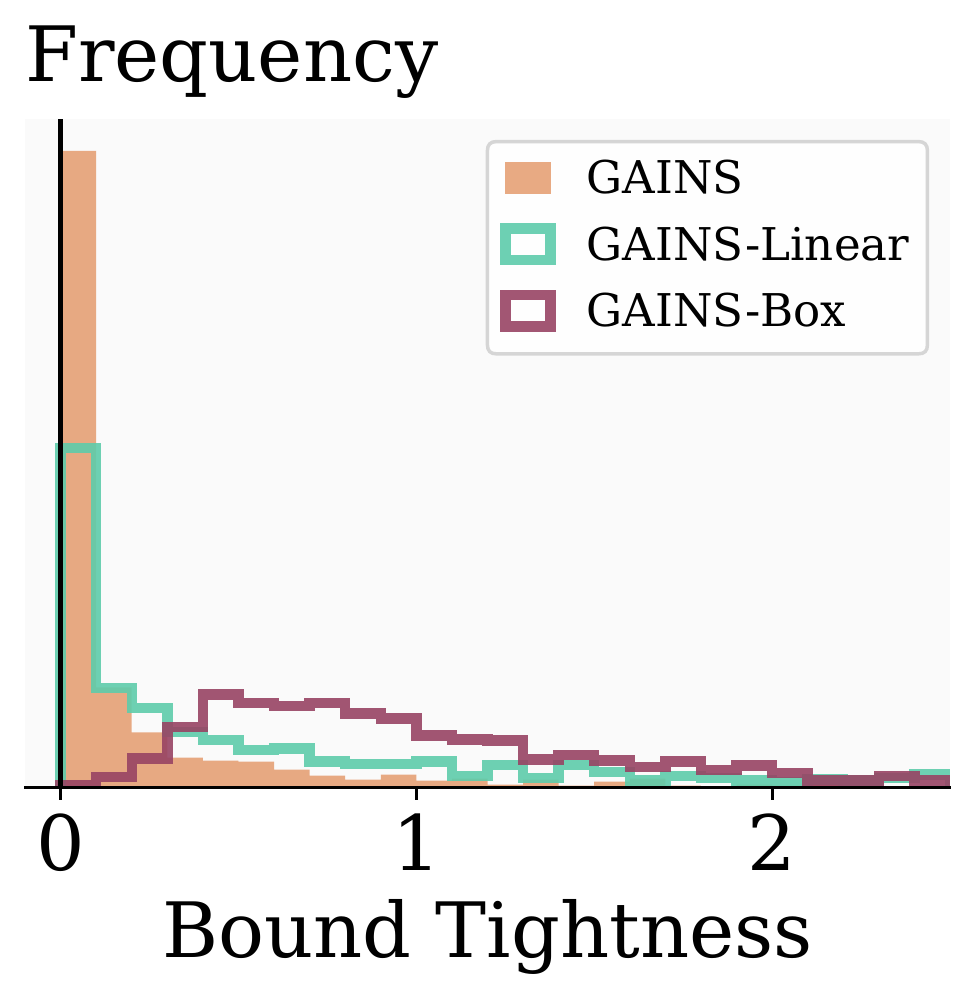}  
	\end{subfigure}
	\caption{Comparison of certified (\tool and \tool-Linear) and empirical (Adv.) bounds on the worst case logit-difference (left) and illustration of the frequency of the bound tightness depending on the verification method (right).}
	\label{fig:tight}
\end{wrapfigure}
In \cref{fig:tight}, we compare the tightness of the certified bounds computed with \tool, \tool-Linear and \tool-\boxd to empirical bounds obtained via an adversarial attack on a \tool-trained NODE ($\epsilon_t=0.16$) for \fmnist. We illustrate both the certified over adversarial bounds (left) and the frequency of different tightness-gap sizes depending on the verification method (right) for a perturbation magnitude of $\epsilon= 0.15$.
In both settings, we evaluate the first 1000 test-set images and compute the empirical bounds with a strong PGD attack using 200 steps. 
We clearly observe that using \tool significantly improves bound tightness.

\subsection{Scalability to \cifar}
In this section, we evaluate the scalability of \tool to the \cifar dataset\citep{krizhevsky2009learning}.
Training a \node with \tool as described below, we report standard, adversarial and certified accuracies in \cref{tab:results_cifar}. We observe that for most perturbation magnitudes ($\epsilon_t = \epsilon = 0.001$), we achieve a standard accuracy of over $60\%$ and and a certified accuracy of $57.1\%$, demonstrating the scalability of our approach to \cifar. 

\begin{table}[t!]
	\centering	
	\caption{The standard (Std.), adversarial (Adv.), and certified (Cert.) accuracy obtained with \tool evaluated on the first 1000 \cifar test set samples .\vspace{-2mm}}
	\begin{threeparttable}
		\renewcommand{\arraystretch}{0.98}
		\begin{tabular}{lccccc}
			\toprule
			\multirow{2.5}*{Training Method} & \multirow{2.5}*{$\epsilon_t$} &  & \multirow{2.5}*{Std. [\%]} & \multicolumn{2}{c}{$\epsilon = 0.001$} \\
			\cmidrule(rl){5-6} 
			&&&& Adv. [\%] & Cert. [\%] \\
			\midrule
			{\toolt} & 0.001    &  & 60.8 &57.6&57.1\\
			\bottomrule
		\end{tabular}

		\label{tab:results_cifar}
	\end{threeparttable}
	\vspace{-3mm}
\end{table}

\paragraph{Experimental Setup} We modify the experimental details from \cref{app:classification-details} such that they are applicable to the \cifar dataset. 
We use $\vmu = [0.4914, 0.4822, 0.4465]$ and $\boldsymbol{\sigma} = [0.2023, 0.1994, 0.2010]$ for standardization. 

During warm-up, we use the scheduler Smooth($\frac{0.1}{255}$, 10, 40) for $\epsilon$-annealing. During the main training, we use $\kappa = 2$ in epochs 1-25, and $\kappa = 4$ in otherwise. For evaluation, we have used a strong PGD attack with 200 steps.

}

\subsection{Hyperparameter Selection}
In this section, we investigate the effects of different hyperparameter selections in provable NODE training, with emphasis on the trajectory exploration and update sampling described in \cref{app:node_trafo_details}. All experiments in this section were conducted on the \fmnist dataset using provable training with $\epsilon_t =0.16$ and the remaining hyperparameters are as described in \cref{app:classification-details}, except when explicitly stated otherwise.

\paragraph{Aggregation Method} 
As described in \cref{app:node_trafo_details} in training we sample $\kappa$ trajectories from the trajectory graph $\mathcal{G}(\mathcal{Z})$ in order to approximate the bounds of the NODE output $\vz(T_{end})$. 
We compare three approaches on how to combine the $\kappa$ trajectories in training, which we call stack, average and worst case. 
The stack approach considers the bounds from each sampled trajectory individually and can be interpreted as increasing the effective batchsize by factor $\kappa$, since we stack all obtained bounds along the batch dimension and propagate the resulting output through the remainder of the architecture.
On the other hand, the average approach uses the mean of all obtained bounds, whereas the worst case approach uses the loosest bounds for each neuron.
The results are reported in \cref{tab:results_aggr_meth}, where we see that the stack approach performs the best. We assume that this follows from the interpretation that this can be seen as increasing the effective batchsize and results in better gradient estimation. On the other hand, using the worst case approach suffers from gradient information loss, due to the usage of the maximum and minimum operations. 

\begin{table}[h]
    \centering	
    \caption{Means and standard deviations of the standard (Std.) and certified (Cert.) accuracy using different aggregation methods evaluated on the first 1000 test set samples of the \fmnist dataset.\vspace{-2mm}}
    \begin{threeparttable}
    	\renewcommand{\arraystretch}{0.98}
    \begin{tabular}{lccc}
		\toprule
		 \multirow{2.5}*{Aggregation Method} &  \multirow{2.5}*{Std. [\%]} & \multicolumn{2}{c}{Cert. [\%]}\\
       	 \cmidrule(rl){3-4} 
       	 && $\epsilon = 0.10$  & $\epsilon = 0.15$ \\
         \midrule

		stack  & \textbf{71.5}$^{\pm 1.7}$&\textbf{61.2}$^{\pm 2.7}$&\textbf{54.8}$^{\pm 4.1}$\\ %
        average  & 71.0$^{\pm 0.4}$&60.0$^{\pm 1.4}$&52.8$^{\pm 0.9}$\\
        worst case  & 69.0$^{\pm 1.5}$&57.9$^{\pm 2.1}$&50.9$^{\pm 2.1}$ \\ %

    \bottomrule
    \end{tabular}
    \label{tab:results_aggr_meth}
    \end{threeparttable}
\vspace{-4mm}
\end{table}

\paragraph{Annealing Process}
In \cref{tab:results_anneal} we evaluate the influence of the used annealing process for the sample probability $q=q_1=q_2$ during training. We observe that when using a fixed sample probability (last two processes in \cref{tab:results_anneal}), \tool achieves higher accuracies when the sampled trajectories are closer to the reference trajectory, i.e. use smaller $q$. 
We hypothesize that the process $\text{Sin}(0.33,0.33,10, 80)$ considers too many trajectories which occur only due to approximation errors in the abstract domain. 
However, we observe the best performance in all settings, when annealing the sampling probability. We assume that staying close to the reference trajectory in the early stages of training stabilizes the network and reduces the number of vertices in the trajectory graph induced by approximation errors.
On the other hand, it is important to refine the bounds in all parts of the trajectory graph, which is why the annealing works best, if in the end we have a uniform distribution, i.e. $q \approx \frac{1}{3}$.
\begin{table}[h]
    \centering	
    \caption{Means and standard deviations of the standard (Std.) and certified (Cert.) accuracy using different annealing processes evaluated on the first 1000 test set samples of the \fmnist dataset.\vspace{-2mm}}
    \begin{threeparttable}
    	\renewcommand{\arraystretch}{0.98}
    \begin{tabular}{lccc}
		\toprule
        \multirow{2.5}*{Annealing Process} &  \multirow{2.5}*{Std. [\%]} & \multicolumn{2}{c}{Cert. [\%]}\\
       	 \cmidrule(rl){3-4} 
       	 && $\epsilon = 0.10$  & $\epsilon = 0.15$ \\
         \midrule
		 $\text{Sin}(0.15,0.33,10, 80)$ & \textbf{71.5}$^{\pm 1.7}$&\textbf{61.2}$^{\pm 2.7}$&\textbf{54.8}$^{\pm 4.1}$\\ 
		 $\text{Sin}(0.15,0.4,10, 80)$  & 68.1$^{\pm 2.6}$&56.4$^{\pm 4.6}$&49.6$^{\pm 5.2}$ \\ 
		 $\text{Sin}(0.15,0.15,10, 80)$  & 70.8$^{\pm 1.2}$&60.1$^{\pm 0.7}$&53.6$^{\pm 0.6}$ \\ 
		 $\text{Sin}(0.33,0.33,10, 80)$  & 68.5$^{\pm 0.7}$&57.8$^{\pm 1.8}$&50.9$^{\pm 2.9}$\\

    \bottomrule
    \end{tabular}
    \label{tab:results_anneal}
    \end{threeparttable}
\vspace{-4mm}
\end{table}

\paragraph{Number of Sampled Trajectories}
In \cref{tab:results_kappa} we evaluate the influence of the number of sampled trajectories $\kappa$, where we additionally investigate the effect of including the reference trajectory among the selected trajectories (\emph{fixed} in \cref{tab:results_kappa}). 
We consider three $\kappa$ settings, in the first one we always use $\kappa = 1$, in the second one we use $\kappa\in[2,4,8]$ as described in \cref{app:classification-details}, and in the last setting, we always use $\kappa = 4$.
We observe that in the $\kappa = 1$ setting it is better to always use the reference trajectory instead of sampling. When increasing $\kappa$, we note that the variant which does not always include the reference trajectory performs better.
\begin{table}[h]
    \centering	
    \caption{Means and standard deviations of the standard (Std.) and certified (Cert.) accuracy using different $\kappa$ evaluated on the first 1000 test set samples of the \fmnist dataset.\vspace{-2mm}}
    \begin{threeparttable}
    	\renewcommand{\arraystretch}{0.98}
    \begin{tabular}{lcccc}
		\toprule
        \multirow{2.5}*{$\kappa$} & \multirow{2.5}*{Selection Method} &  \multirow{2.5}*{Std. [\%]} & \multicolumn{2}{c}{Cert. [\%]}\\
       	 \cmidrule(rl){4-5} 
       	 &&& $\epsilon = 0.10$  & $\epsilon = 0.15$ \\
         \midrule

        \multirow{2}*{1}& sample $\kappa$   & 69.0$^{\pm 1.6}$&57.8$^{\pm 3.0}$&51.0$^{\pm 3.7}$\\
        & fixed  & 71.5$^{\pm 1.5}$&60.7$^{\pm 1.2}$&54.1$^{\pm 1.0}$\\ 
         \cmidrule(rl){1-2}
         \multirow{2}*{[2,4,8]}& sample $\kappa$  & 71.5$^{\pm 1.7}$&61.2$^{\pm 2.7}$&\textbf{54.8}$^{\pm 4.1}$\\ 
        & fixed + sample $\kappa - 1$   & 70.7$^{\pm 1.5}$&60.2$^{\pm 1.8}$&53.2$^{\pm 1.6}$\\ 
		\cmidrule(rl){1-2}
		\multirow{2}*{4}&  sample $\kappa$    & \textbf{71.8}$^{\pm 0.9}$&\textbf{62.2}$^{\pm 1.0}$&54.7$^{\pm 1.5}$\\
        & fixed + sample $\kappa - 1$  & 69.9$^{\pm 2.0}$&59.2$^{\pm 2.6}$&53.4$^{\pm 3.9}$\\

    \bottomrule
    \end{tabular}
    \label{tab:results_kappa}
    \end{threeparttable}
\vspace{-4mm}
\end{table}

\fi

\end{document}